\documentclass[sigconf]{acmart}

\usepackage{multirow}
\usepackage{cleveref}
\usepackage{float}
\usepackage{subfig}
\usepackage{makecell}
\usepackage{listings}
\usepackage{xcolor}
\usepackage{algorithm}
\usepackage[noend]{algpseudocode}
\usepackage{enumitem}
\usepackage{adjustbox}

\newtheorem*{theorem*}{Theorem}
\newtheorem*{remark*}{Remark}

\DeclareMathOperator*{\argmin}{argmin}

\algnewcommand\INPUT{\item[{\textbf{Input:}}]}
\algnewcommand\PREPROCESS{\item[{\textbf{Preprocess:}}]}
\algnewcommand\TRAINING{\item[{\textbf{Training:}}]}
\algnewcommand\INFERENCE{\item[{\textbf{Inference:}}]}
\algnewcommand\OUTPUT{\item[{\textbf{Output:}}]}

\lstdefinestyle{mystyle}{
    language=Python,
    basicstyle=\fontsize{8.5pt}{10pt}\selectfont\ttfamily,
    frame=single,
    framesep=2.5pt,
    breaklines=true,
    keywordstyle=\color{blue},
    commentstyle=\color{gray},
    stringstyle=\color{red},
    numberstyle=\tiny,
    numbers=none,
    showstringspaces=false,
    tabsize=4,
    columns=flexible,
}

\begin{document}

\title{Interface Laplace Learning: Learnable Interface Term Helps Semi-Supervised Learning}

\author{Tangjun Wang}
\affiliation{%
  \institution{Tsinghua University}
  \state{Beijing}
  \country{China}}
\email{wangtj20@mails.tsinghua.edu.cn}

\author{Chenglong Bao}
\authornote{Corresponding authors}
\affiliation{%
  \institution{Tsinghua University}
  \state{Beijing}
  \country{China}}
\email{clbao@mail.tsinghua.edu.cn}

\author{Zuoqiang Shi}
\authornotemark[1]
\affiliation{%
  \institution{Tsinghua University}
  \state{Beijing}
  \country{China}}
\email{zqshi@tsinghua.edu.cn}

\begin{abstract}
We introduce a novel framework, called Interface Laplace learning, for graph-based semi-supervised learning. Motivated by the observation that an interface should exist between different classes where the function value is non-smooth, we introduce a Laplace learning model that incorporates an interface term. This model challenges the long-standing assumption that functions are smooth at all unlabeled points. In the proposed approach, we add an interface term to the Laplace learning model at the interface positions. We provide a practical algorithm to approximate the interface positions using k-hop neighborhood indices, and to learn the interface term from labeled data without artificial design. Our method is efficient and effective, and we present extensive experiments demonstrating that Interface Laplace learning achieves better performance than other recent semi-supervised learning approaches at extremely low label rates on the MNIST, FashionMNIST, and CIFAR-10 datasets.
\end{abstract}




\maketitle

\section{Introduction}

The success of machine learning methods often depends on a large amount of training data. However, collecting training data can be labor-intensive, and is sometimes impossible in many application fields due to privacy or safety issues. To alleviate the dependency on training data, semi-supervised learning (SSL)~\cite{zhu2009introduction,chapelle2009semi} has received great interest in recent years. Semi-supervised learning typically uses a large amount of unlabeled data, together with the labeled data, to improve model performance and generalization ability. The idea of combining labeled and unlabeled data has been widely used long before the term SSL was coined~\cite{fralick1967learning, blum1998combining}. By incorporating the geometric structure or data distribution of unlabeled data, SSL algorithms aim to extract more informative features, thereby enhancing the performance of machine learning models.

This paper focuses on a type of SSL method: graph-based SSL~\cite{song2022graph}. Graph-based SSL algorithms have received much attention as the graph structure can effectively encode relationships among data points, thereby allowing for full utilization of the information contained in unlabeled data. Graph-based SSL is based on the assumption that nearby nodes tend to have the same labels. In a graph, each sample is represented by a vertex, and the weighted edge measures the similarity between samples. One of the most widely used methods in Graph-based SSL is the Gaussian Fields and Harmonic Functions algorithm~\cite{zhu2003semi}, later commonly called Laplace learning. Laplace learning aims to minimize the graph Dirichlet energy with the constraint on labeled points, resulting in a harmonic function. Many variants of Laplace learning have been proposed. One way is to replace the hard label constraint with soft label regularization~\cite{zhou2004learning,belkin2004regularization,kang2006correlated,ando2007learning}. Another way is to generalize the $\ell^2$ distance, which corresponds to Laplace learning, into $\ell^p$ distance, known as $p$-Laplace learning~\cite{zhou2005regularization,buhler2009spectral,bridle2013p,el2016asymptotic, slepcev2019analysis}. In the limit as $p$ approaches infinity, $p$-Laplace learning is called Lipschitz learning~\cite{kyng2015algorithms}.

However, it has been observed that Laplace learning and its variants exhibit poor performance when the label rate is low~\cite{nadler2009semi,el2016asymptotic,calder2020poisson}. In these situations, the solutions tend to converge to non-informative, nearly constant functions with spikes near labeled points, significantly deteriorating the model's accuracy. To address the issue, several methods have been proposed, e.g. higher-order regularization~\cite{zhou2011semi}, graph re-weighting~\cite{shi2017weighted,calder2020properly}, spectral cutoff~\cite{belkin2002using} and centered kernel~\cite{mai2018random}. While these methods have been explored, they are either much more computationally burdensome, or still perform poorly when the label rate is extremely low. Recently, Poisson learning~\cite{calder2020poisson} has shown promising results in this challenging scenario. Poisson learning replaces the Dirichlet boundary conditions in the Laplace equation with a source term in the Poisson equation, achieving a significant performance increase in classification tasks under extreme label rates.

Nonetheless, most existing methods assume that the solution should exhibit a certain degree of smoothness across all unlabeled points. In this work, we demonstrate that, ideally, there should exist an \textit{interface} between two different classes. The solution should exhibit discontinuity on the interface, rather than being globally smooth. Interface problems are commonly encountered in fields like materials science~\cite{emmerich2003diffuse}, fluid dynamics~\cite{shyy1999fluid,constantin2001some}, and electromagnetic~\cite{mahan1956reflection,stephan1983solution}, where the solution is expected to have clear boundaries between different regions or classes, rather than a smooth transition.

In this paper, we formulate the interface problem in graph-based SSL as solving the Laplace equation with jump discontinuity across the interface. By deriving the nonlocal counterpart of this system, we find that it is naturally related to Laplace learning, but with the addition of an explicit interface term. By accounting for the non-smoothness across the interface, our approach can more accurately model the underlying data distribution, resulting in improved performance compared to standard Laplace learning and its variants.

Our main contribution can be summarized as follows:
\begin{itemize}[leftmargin=*]
    \item We are the first to propose the concept of interface discontinuity in graph-based SSL, offering a new perspective and algorithmic design in this domain.
    \item We introduce a Laplace learning model that explicitly accounts for interface discontinuities, improving upon existing approaches.
    \item We develop a practical algorithm to approximate the interface and learn the interface term without deliberate design.
    \item On benchmark classification tasks with extremely low label rates, our method achieves state-of-the-art performance.
\end{itemize}

\section{Motivation}
\subsection{Laplace Learning and Poisson Learning}
SSL aims to infer the labels of unlabeled samples $\{x_{m+1}, \cdots, x_{n}\}$ with the help of a labeled set, $\{(x_1,y_1), \cdots,\allowbreak (x_m,y_m)\}$. For a classification problem with $c$ classes, the label $y_i$ is often represented as a one-hot vector in $\mathbb{R}^c$, where the $l$-th element is 1 if the sample belongs to class $l$, and the other elements are all zeros. Typically, the number of all samples $n$ is much larger than the number of labeled samples $m$. Graph-based SSL methods construct a graph where the nodes correspond to the samples, $V=\{x_1,x_2, \cdots, x_n\}$. An edge $e_{ij}$ is created between nodes $x_i$ and $x_j$ if the two samples are similar, and a non-negative weight $w_{ij}$ is assigned to the edge, indicating the degree of similarity between $x_i$ and $x_j$.

Among numerous graph-based SSL approaches, Laplace learning~\cite{zhu2003semi}, also known as label propagation, is one of the most widely used methods. Laplace learning propagates labels to unlabeled nodes by solving the following Laplace equation:
\begin{alignat*}{2}
    Lu\left(x_i\right)&=0 && \quad m+1 \leq i \leq n \\
	u\left(x_i\right)&=y_i && \quad 1 \leq i \leq m
\end{alignat*}
where $L$ is the graph Laplacian operator given by
\[L u\left(x_i\right) =\sum_{j=1}^n w_{ij} (u(x_i)-u(x_j))\]
The solution $u:V\rightarrow \mathbb{R}^c$ gives the prediction of size $c$ for each vertex $x_i$. The label decision is then determined by the largest component in $u(x_i)$. In Laplace learning, labeled data is incorporated as Dirichlet boundary conditions $u\left(x_i\right)=y_i$.

Recently, Poisson learning~\cite{calder2020poisson} is proposed as an alternative to Laplace learning to deal with extremely low label rate. Poisson learning treats unlabeled data in the same way as Laplace Learning, but it differs in the way of handling labeled data. Poisson learning replaces the Dirichlet boundary condition with the Poisson equation and a given source term $y_i-\bar{y}$,
\begin{alignat*}{2}
    & Lu\left(x_i\right)=0 && \quad m+1 \leq i \leq n \\
	& Lu\left(x_i\right)=y_i-\bar{y} && \quad 1 \leq i \leq m
\end{alignat*}
where $\bar{y}=\frac{1}{m}\sum_{j=1}^{m}y_j$. Surprisingly, such modification can result in huge improvement under extremely low label rates, e.g. 16.73\% $\rightarrow$ 90.58\% in MNIST~\cite{lecun1998gradient} 1-label per class classification.

\subsection{Interface discontinuity}
\label{subsec:interface_discontinuity}

Both Laplace learning and Poisson learning assume that the graph Laplacian of the target function $Lu(x_i)=0$ on all unlabeled points. However, we question whether this is the most appropriate way to model the underlying data distribution. To illustrate our idea, we will consider a synthetic 2-class classification example.

\begin{figure}[hbtp]
    \centering
    \subfloat[Ground Truth]{\includegraphics[width=0.4\linewidth]{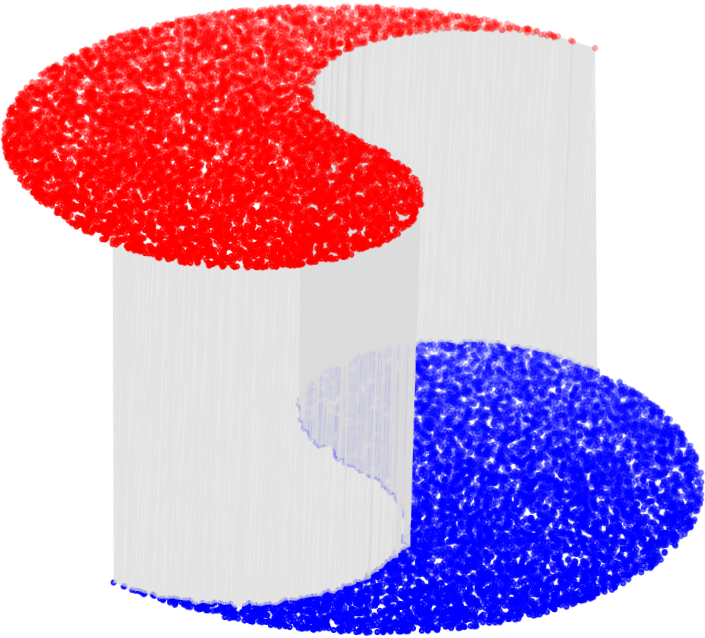}}
    \hspace{0.1\linewidth}
    \subfloat[$L u\left(x_i\right)\neq 0$ on the interface]{\includegraphics[width=0.4\linewidth]{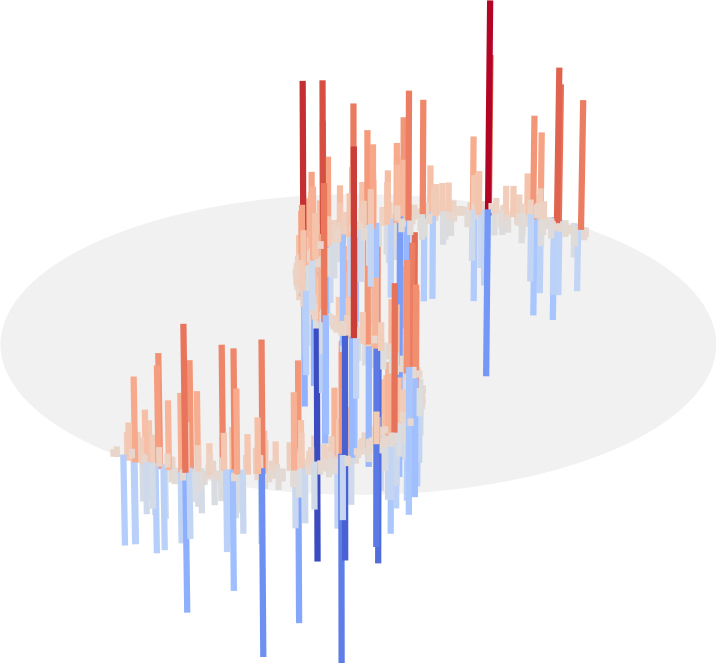}}
    
    \caption{3D visualization of a synthetic classification example.}
    \Description{A synthetic 2-class classification example.}
    \label{fig:classification_3d}
\end{figure}

We uniformly sample 20,000 points $x_i=(a_i,b_i)\in \mathbb{R}^2$ from a unit circle. The decision boundary between the two classes is defined as the union of two half-circles: $\{a < 0, a^2+(b-0.5)^2=0.5^2 \}\cup \{a \geq 0, a^2+(b+0.5)^2=0.5^2\}$. The ground truth label for each $x_i$ is in $\{-1,+1\}\in \mathbb{R}$. A 3D visualization of this toy example is given in~\Cref{fig:classification_3d}(a). To construct the similarity matrix $\mathbf{W}=(w_{ij})\in \mathbb{R}^{n\times n }$, we use a Gaussian kernel defined as:
\begin{equation}
    w_{ij}=\exp\left( -\frac{4\|x_i-x_j\|^2}{d_K(x_i)^2}\right)
    \label{eq:gaussian_kernel}
\end{equation}
where $d_K(x_i)$ is the distance between $x_i$ and its $K$-th nearest neighbor. We choose $K=10$, and we sparsify the matrix $\mathbf{W}$ by truncating the weight for points farther than the $K$-th nearest neighbor to zero.

We will examine the problem from two different perspectives. Firstly, we assign the ground truth labels to the function $u(x_i)$ and then compute and plot $Lu(x_i)\in\mathbb{R}$ in~\Cref{fig:classification_3d}(b). From the figure, it is clear that the Laplacian of the labeling function is nonzero along the interface between the two classes, while being zero in the interior of each class. This nonzero Laplacian near the decision boundary is due to the interface discontinuity in the labeling function. However, this interface discontinuity is ignored by both Laplace learning and Poisson learning, as they assume the function is harmonic almost everywhere, except at the labeled points. The assumption of harmonicity contradicts the true Laplacian behavior observed in the ground truth labeling function.


\begin{figure}[hbtp]
    \centering
    \captionsetup[subfigure]{justification=centering}
    \subfloat[Ground Truth]{\includegraphics[width=0.25\linewidth]{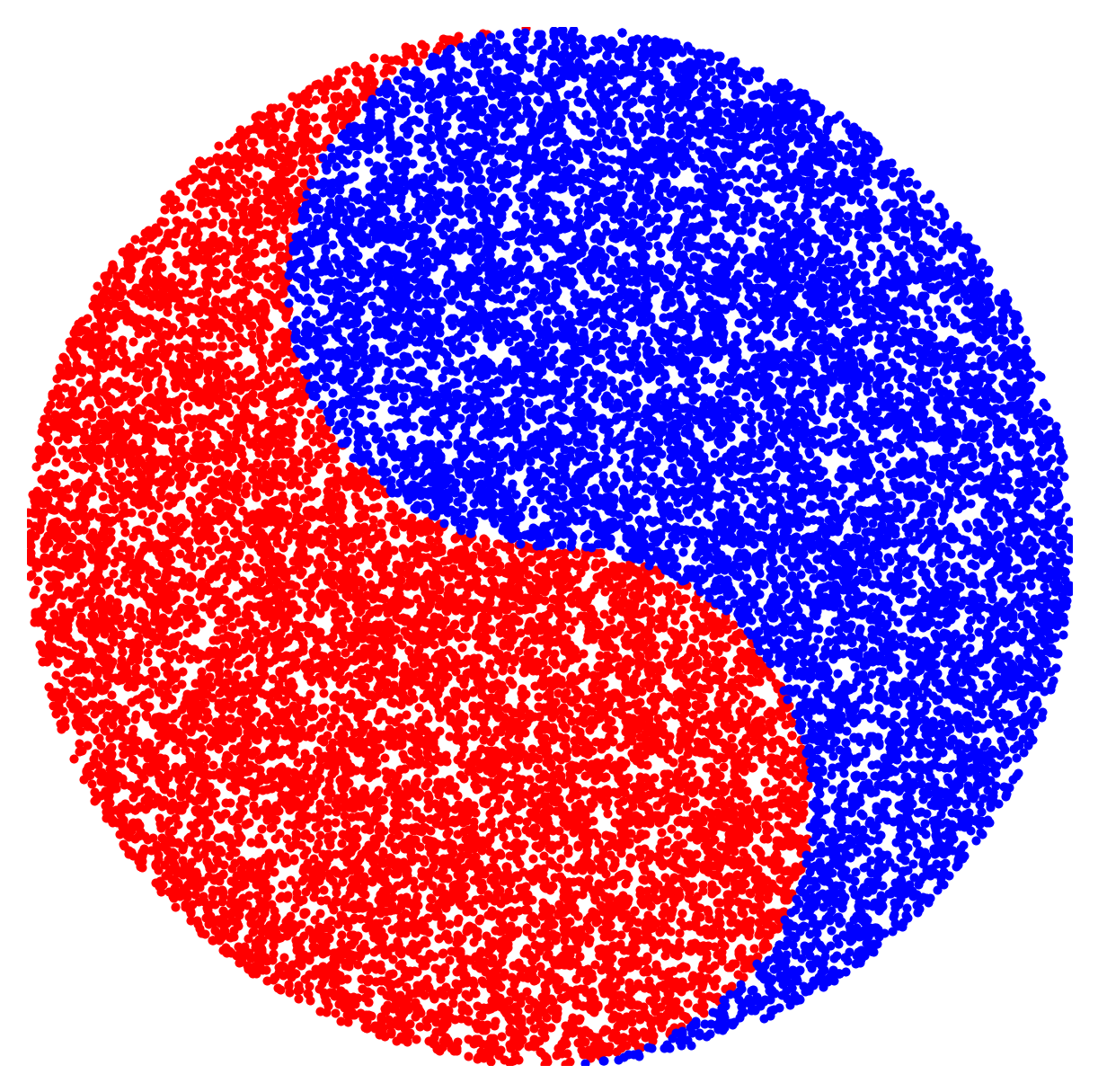}}
    \subfloat[Laplace \\ Acc=93.87\%]{\includegraphics[width=0.25\linewidth]{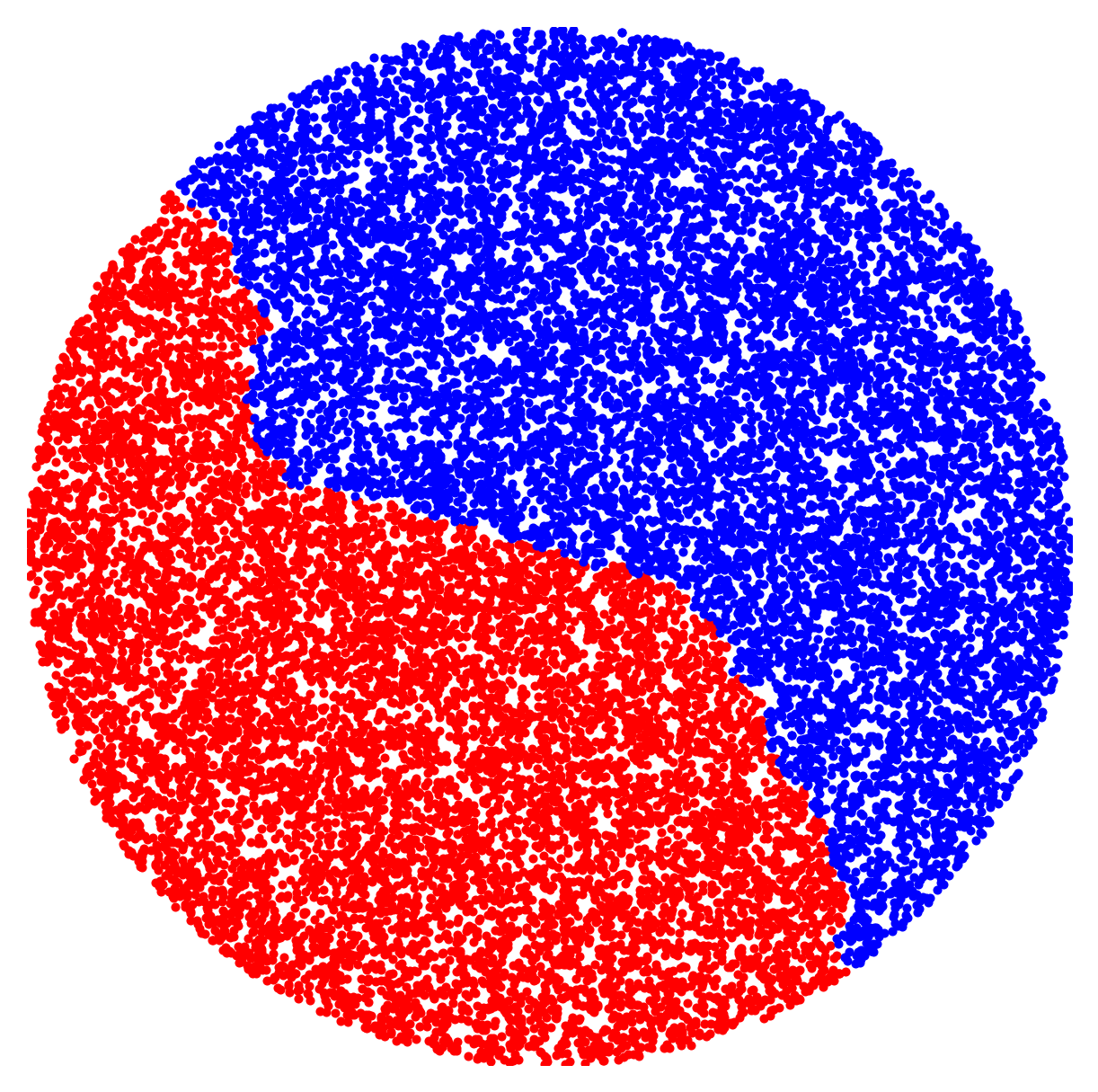}}
    \subfloat[Poisson \\Acc=91.17\%]{\includegraphics[width=0.25\linewidth]{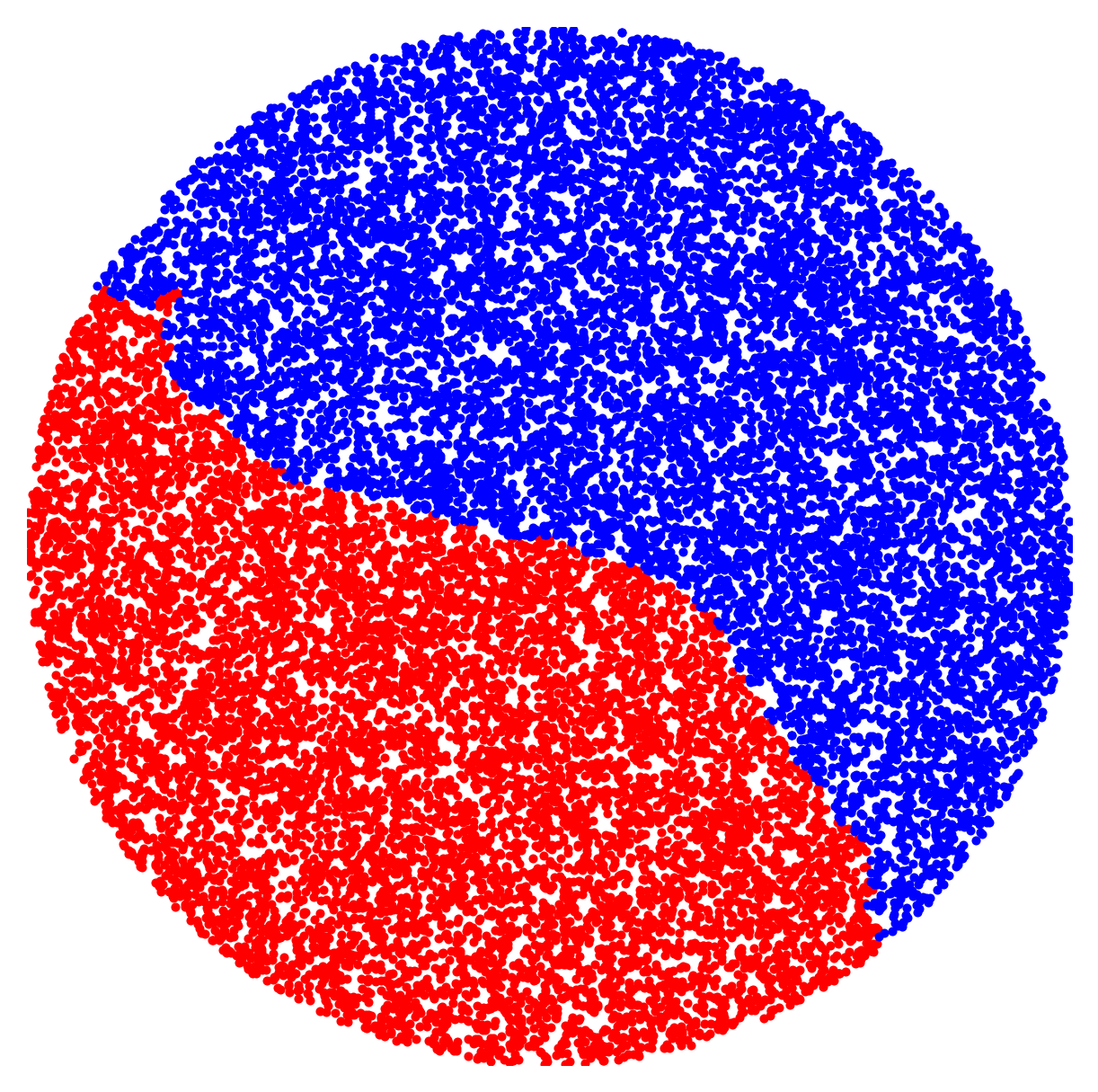}}
    \subfloat[Ours \\Acc=98.43\%]{\includegraphics[width=0.25\linewidth]{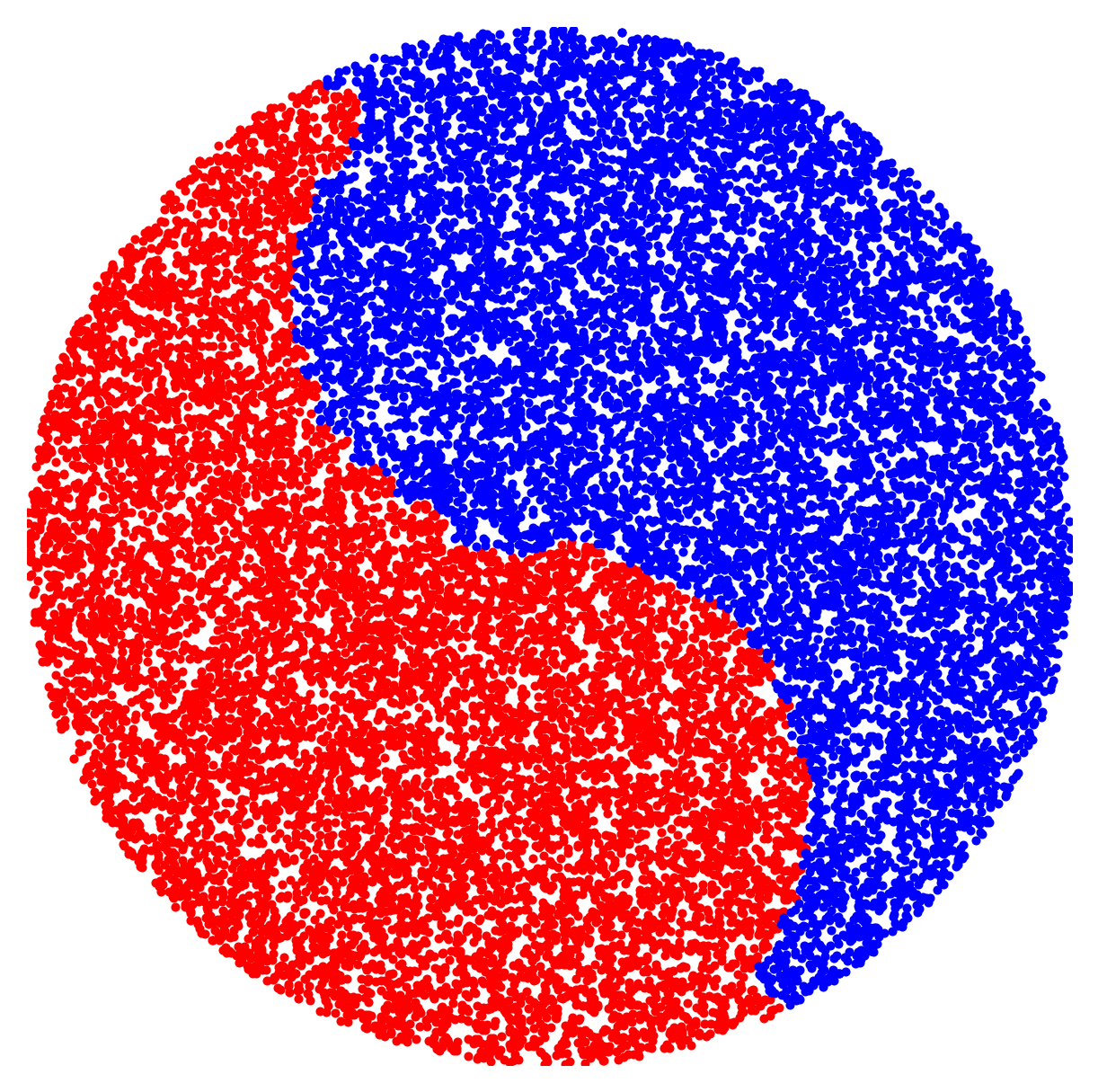}}
    \caption{2D visualization of a synthetic classification example.}
    \Description{A synthetic 2-class classification example.}
    \label{fig:classification}
\end{figure}

Secondly, we will show that adding an interface term to account for the interface discontinuity is helpful for classification. Specifically, we choose to modify the Laplace equation by introducing an interface term $f_i$ that will be learned from the labeled data:
\begin{equation}
    \begin{alignedat}{2}
        & Lu\left(x_i\right)=0 && \quad i \notin \mathcal{I} \\
        & Lu\left(x_i\right)=f_i && \quad i \in \mathcal{I}
    \end{alignedat}
    \label{eq:interface_laplace}
\end{equation}
Here, $\mathcal{I}$ denotes the set of indices corresponding to the interface positions. This formulation allows us to explicitly model the nonzero Laplacian at the interface, in contrast to the assumptions made by standard Laplace and Poisson learning methods. 

The theoretical basis for introducing this interface term to account for interface discontinuity will be provided in the next subsection. In this toy example, the set $\mathcal{I}$ is obtained by identifying the indices where the ground truth Laplacian, as shown in \Cref{fig:classification_3d}(b), is nonzero. The values of $f_i$ will then be inferred from the labeled points using the algorithm detailed in~\Cref{sec:method}. Finally, we use the solution $u$ to Eq.~\eqref{eq:interface_laplace} for classification.

We randomly select 25 labeled samples from each class, and use Laplace learning, Poisson learning, and our proposed method to classify the remaining data points. The classification results are provided in \Cref{fig:classification}. We can observe that our method significantly outperforms both Laplace learning and Poisson learning in terms of classification accuracy. Moreover, the decision boundary obtained by our method is also much closer to the ground truth. Indeed, the comparison is unfair because our method utilizes additional information about the interface positions, which the other two methods do not have access to. However, this toy classification problem serves to verify our argument that incorporating an interface term to account for the discontinuity is both necessary and beneficial for improving classification performance.

\subsection{Laplace equation with interface and associate nonlocal model}
In this subsection, we will provide a theoretical verification of our approach from the perspective of nonlocal models. Nonlocal models~\cite{du2019nonlocal} play a crucial role in many fields, such as peridynamical theory of continuum mechanics, nonlocal wave propagation and nonlocal diffusion process~\cite{alfaro2017propagation,bavzant2003nonlocal,blandin2016well,dayal2007real,kao2010random,vazquez2012nonlinear}. The terminology \textit{nonlocal} is the counterpart of \textit{local} operators. A linear operator $L$ is considered local if the support set $\text{supp}\{L(f)\} \subset \text{supp}\{f\}$ for any function $f$. Differential operators, such as the Laplace operator $\Delta$, are local operators. 

As discussed in previous subsection, the interface problem in SSL can be modeled by the following Laplace equations with jump discontinuity on the interface $\Gamma$,
\begin{equation}
    \begin{alignedat}{2}
        \Delta u^+(x)&=0 && \quad x\in \mathcal{M}_1 \\
        \Delta u^-(x)&=0 && \quad x\in \mathcal{M}_2 \\
        u^+(x)-u^-(x)&=1 && \quad x\in \Gamma\\
        \frac{\partial u}{\partial \mathbf{n}}^+(x)-\frac{\partial u}{\partial \mathbf{n}}^-(x)&=0 && \quad x\in \Gamma      
    \end{alignedat}
    \label{eq:local_model}
\end{equation}
Here $\Gamma$ is a $(k-1)$-dimensional smooth manifold that splits a $k$-dimensional manifold $\mathcal{M}$ into two submanifolds $\mathcal{M}_1$ and $\mathcal{M}_2$, so that $\mathcal{M}_1\cap\mathcal{M}_2=\emptyset$ and $\mathcal{M}=\mathcal{M}_1\cup\Gamma\cup\mathcal{M}_2$. $\mathbf{n}$ is the outer normal of $\Gamma$. 

To get nonlocal approximation, we introduce a rescaled kernel function $R_{\delta}(x,y)=C_\delta R\big(\frac{\|x-y\|^2}{4\delta^2}\big)$, where $R\in C^2([0,1])$ is a non-negative compact function that is supported over $[0,1]$. $C_{\delta }=( 4\pi \delta ^2)^{-d/2}$ is a normalization factor for $x\in \mathbb{R}^d$. The rescaled functions $\bar{R}_{\delta}(x,y)$ and $\bar{\bar{R}}_{\delta}(x,y)$ are defined similarly, where $\bar{R}(r)=\int_{r}^{+ \infty }R( s) ds$ and $\bar{\bar{R}}(r)=\int_{r}^{+ \infty }\bar{R}( s) ds$. The role of kernel function is similar to the similarity matrix $\mathbf{W}$ in Eq.~\eqref{eq:gaussian_kernel}, where only nearest neighbors have nonzero weights. If we assume $\int_{\mathcal{M}_{1}} \bar{R}_{\delta}(y,s)d\mu_{s}=\int_{\mathcal{M}_{2}} \bar{R}_{\delta}(y,s)d\mu_{s}:= w_\delta(y)$, then \cite{zhang2021nonlocal} gives a nonlocal model to approximate Eq.~\eqref{eq:local_model} as:
\begin{align}
    \begin{split}
        \int_{\mathcal{M}_1}R_\delta(x,y)(u_\delta^+(x)-u_\delta^+(y))d\mu_y = \int_{\Gamma} \bar{R}_\delta(x,y) v_\delta(y) d\tau_y \hspace{17.5pt} & \\
        x\in \mathcal{M}_1 \\
        \int_{\mathcal{M}_2}R_\delta(x,y)(u_\delta^-(x)-u_\delta^-(y))d\mu_y = - \int_{\Gamma} \bar{R}_\delta(x,y) v_\delta(y) d\tau_y \hspace{8pt} & \\
        x\in \mathcal{M}_2 \\
    \end{split}
    \label{eq:nonlocal_model}
\end{align}
where 
\[v_\delta(y)=\frac{w_\delta(y) + \int_{\mathcal{M}_{2}} u_\delta^-(s) \bar{R}_{\delta}(y,s)d\mu_{s} - \int_{\mathcal{M}_{1}} u_\delta^+(s) \bar{R}_{\delta}(y,s)d\mu_{s}}{2\int_{\Gamma} \bar{\bar{R}}_{\delta}(y,s)d\tau_{s}}\]
is an approximation to the normal derivative up to a constant.


 The solutions of the nonlocal model are proven to converge to the solutions of the local model with the rate of $O(\delta)$ in $H^1$ norm.
\begin{theorem*}[~\cite{zhang2021nonlocal}]
    (1) (Well-Posedness) \hspace{0.5em} For any $f\in L^2(\mathcal{M})$, there exists a unique solution $(u_\delta^+, u_\delta^-) \in (H^1(\mathcal{M}_1),H^1(\mathcal{M}_2))$ to Eq.~\eqref{eq:nonlocal_model}. \quad (2) (Convergence) \hspace{0.5em} For any $f\in H^1(\mathcal{M}_1 \cup \mathcal{M}_2)$, let $(u^+, u^-) \in \left(H^3(\mathcal{M}_1),\allowbreak H^3(\mathcal{M}_2)\right)$ be the solution to Eq.~\eqref{eq:local_model}, and $(u_\delta^+, u_\delta^-)$ be the solution to Eq.~\eqref{eq:nonlocal_model}, then
    \begin{align*}
        & \left\|u^+-u_\delta^{+}\right\|_{H^1(\mathcal{M}_1)}+\left\|u^{-}-u_\delta^-\right\|_{H^1(\mathcal{M}_2)} \\
        & \hspace{5em} \leq C\delta \left( \left\|u^+\right\|_{H^3(\mathcal{M}_1)} + \left\|u^-\right\|_{H^3(\mathcal{M}_2)}\right)
    \end{align*}
    where the constant C only depends on $\mathcal{M}$ and $\Gamma$.
\end{theorem*}

Nonlocal model formulation \eqref{eq:nonlocal_model} provides theoretical justification of incorporating an interface term to account for interface discontinuity in our algorithm \eqref{eq:interface_laplace}. On the left-hand side of the nonlocal model, We can identify the integral $\int_{\mathcal{M}_1}R_\delta(x,y)(u^+_\delta(x)-u^+_\delta(y))d\mu_y$ with the graph Laplacian operator $L u\left(x_i\right) =\sum_{j=1}^n w_{ij} (u(x_i)-u(x_j))$ by discretizing the integral. On the right-hand side, since $\bar{R}_\delta(x,y)$ has compact support, the integral $\int_{\Gamma} \bar{R}_\delta(x,y) v_\delta(y) d\tau_y$ is nonzero only when $x$ lies in a layer adjacent to the interface $\Gamma$ with width $2\delta$. We can relate this adjacent layer of $\Gamma$ to the interface positions $\mathcal{I}$ introduced earlier. This explains our choice of introducing $f_i=Lu(x_i)\neq 0$ for $i\in \mathcal{I}$.

While the nonlocal formulation provides theoretical justification for the interface term, it is not directly applicable to semi-supervised learning. The variables $v_\delta$ and $(u_\delta^+,u_\delta^-)$ are coupled together, making it challenging to write Eq.~\eqref{eq:nonlocal_model} into a linear system. Furthermore, the interface $\Gamma$ is not given explicitly, which makes the computation of the integral over $\Gamma$ impossible. To address these limitations, we will make the interface terms learnable and provide a practical algorithm in subsequent section. 

\begin{remark*}
    The assumptions $\frac{\partial u}{\partial n}^+(x)=\frac{\partial u}{\partial n}^-(x)$ for $x \in \Gamma$ and $\int_{\mathcal{M}_{1}} \bar{R}_{\delta}(y,s)d\mu_{s}=\int_{\mathcal{M}_{2}} \bar{R}_{\delta}(y,s)d\mu_{s}$ are introduced to simplify the expression of the nonlocal model Eq.~\eqref{eq:nonlocal_model}. However, these assumptions can be relaxed by introducing coefficients: $\lambda_1\frac{\partial u}{\partial n}^+(x)=\lambda_2\frac{\partial u}{\partial n}^-(x)$, $\gamma_\delta(y)=\int_{\mathcal{M}_{2}} \bar{R}_{\delta}(y,s)d\mu_{s}/\int_{\mathcal{M}_{1}} \bar{R}_{\delta}(y,s)d\mu_{s}$. Correspondingly, the nonlocal model should be slightly modified to incorporate these coefficients. The exact form of the modified nonlocal model is provided in~\cite{zhang2021nonlocal}.
\end{remark*}

\section{Method}
\label{sec:method}
In the previous section, from the premise that the function $u$ should be discontinuous at the interface, we propose a formal algorithm:
\begin{alignat*}{2}
    & Lu\left(x_i\right)=0 && \quad i \notin \mathcal{I} \\
    & Lu\left(x_i\right)=f_i && \quad i \in \mathcal{I}
\end{alignat*}

However, two key questions remain unsolved:
\begin{itemize}[leftmargin=*]
    \item For $i\in \mathcal{I}$, how do we decide the value of interface term $f$?
    \item For general classification problems where the interface is not provided a priori, how do we determine $\mathcal{I}$?
\end{itemize}

In the subsequent two subsections, we will provide ideas to address these questions, and the final algorithm will be presented in \Cref{subsec:algorithm}.

\subsection{Interface term learning}
The idea is straightforward: we aim to learn an interface term such that the solution to the modified Laplace equation at the labeled points closely matches the given labels. To this end, we use the Mean Squared Error~(MSE) on the labeled points as the objective function:
\[ \mathcal{L}_{\text{MSE}}=\frac{1}{m}\sum_{i=1}^{m} \|u(x_i)-y_i\|_2^2\]

In addition to the MSE loss, we find that a regularizer on the norm of $f_i$ is beneficial for learning:
\[\mathcal{L}_{\text{reg}}=\sum_{i=1}^{n}\|f_i\|_2^2\]
The introduction of this regularization term is essential, as it (1) ensures the uniqueness of the minimizer of the objective function; (2) facilitates computational efficiency by justifying the later use of the Sherman–Morrison–Woodbury formula; and (3) suppresses the magnitude of the interface term, thereby preventing overfitting. The final objective function is 
\[\mathcal{L} =\mathcal{L}_{\text{MSE}} + \lambda \mathcal{L}_{\text{reg}}\]
where $\lambda$ is a weighting factor, later referred to as the ridge parameter. A method for selecting an appropriate $\lambda$, based on the value of $\mathcal{L}_{\text{MSE}}$, will be presented in~\Cref{subsec:algorithm}.

By minimizing this objective function, we can learn the interface term from label information. We will formulate the problem as a standard least squares~(LS) problem, and the detailed optimization algorithm will be provided in~\Cref{subsec:algorithm}.

\subsection{Interface positions approximation}

In the synthetic 2-class classification example presented in \Cref{subsec:interface_discontinuity}, the interface positions $\mathcal{I}$ are obtained by leaking label information and given as an additional input. However, in real-world classification tasks, the interface positions are generally not known a priori. Furthermore, compared to the two-dimensional and well-separated synthetic data, real-world datasets often possess high-dimensional data with much more sophisticated data distributions. Even if the interface between categories exists, it is impossible for us to know the exact locations of these interfaces.

Here, We approximate the interface positions by excluding the k-hop neighbors of training samples, where k-hop neighbors of a node $v$ are those within distance $k$ from $v$ in the graph. The distance between two vertices in a graph is defined as the number of edges in a shortest path connecting them. We efficiently obtain the k-hop neighbors using an iterative approach, starting with the training samples (distance-0) and finding their direct neighbors (distance-1), then their neighbors (distance-2), and so on. We provide the Python implementation of this get\_interface\_idx() method in the following.

\begin{lstlisting}
# train_idx : numpy array, shape=[m]
# all_idx   : numpy array, shape=[n]
# W         : scipy sparse matrix, shape=[n, n]
# k         : k hop

import numpy as np

def get_interface_idx(train_idx, all_idx, W, k):
    if k == -1:
        return all_idx
    else:
        khop_idx = train_idx
        for _ in range(k):
            neighbor_idx = W[khop_idx].nonzero()[1]
            khop_idx = np.append(khop_idx, neighbor_idx)
            khop_idx = np.unique(khop_idx)
        interface_idx = np.setdiff1d(all_idx, khop_idx)
    return interface_idx
\end{lstlisting}

The reason we approximate the interface positions in this way is two-fold. First, our approach is based on two key assumptions: (1) $\mathcal{I}$ should lie near the class boundaries; and (2) the labeled points should be \emph{representative} of their respective classes—i.e., located well within the interior of each class. The second assumption is particularly important in the extremely low label rate regime: if there is only one labeled point per class and it is not representative, good classification performance is unlikely. Based on these assumptions, it is natural to identify the interface points as those far from the labeled ones. Second, choosing $\mathcal{I}$ away from labeled points is also helpful to avoid overfitting. Empirically, when learning the interface term from extremely limited labeled data, directly adjusting $f_i$ for indices $i$ corresponding to labeled samples or their nearby neighbors can have a disproportionately large effect on minimizing the objective. As a result, such easily adjustable components are more prone to overfitting. A discussion of other possible approaches to approximate interface positions will be provided in~\Cref{subsec:interface_index}.

\subsection{Algorithm}
\label{subsec:algorithm}

For ease of writing, we introduce the following notations. $\mathbf{f}=[f_1, \cdots, f_n]^\top$ and $\mathbf{u}=[u(x_1), \allowbreak \cdots, u(x_n)]^\top$ both belong to $\mathbb{R}^{n\times c}$. $\mathbf{y}=[y_1,\allowbreak \cdots, y_m]^\top \in \mathbb{R}^{m\times c}$. The graph Laplacian matrix $\mathbf{L} = \mathbf{D}-\mathbf{W} \in \mathbb{R}^{n\times n}$, where $\mathbf{D} = \text{diag}(d_i)$, $d_i=\sum_{j=1}^n w_{ij}$. $\mathbf{L}$ is related to the graph Laplacian operator $L$ in that $\mathbf{L}\mathbf{u}=[Lu(x_1), \cdots, Lu(x_n)]^\top$. $\|\cdot \|_{F}$ denotes the Frobenius norm of a matrix.

The objective is to solve the following optimization problem after we obtain the interface positions $\mathcal{I}$ using get\_interface\_idx().
\begin{align}
    \begin{split}
        \argmin_{\mathbf{f}} & ~ \frac{1}{m} \sum_{i=1}^{m} \|u(x_i)-y_i\|_2^2 +\lambda \sum_{i=1}^{n}\|f_i\|_2^2 \\
        \text{s.t. } & ~ f_i=0, ~i\notin \mathcal{I}
    \end{split}
    \label{eq:optimization}
\end{align}
$\mathbf{u}$ and $\mathbf{f}$ are related by the Poisson equation $\mathbf{L}\mathbf{u}=\mathbf{f}$. Since we cannot directly write $\mathbf{u}=\mathbf{L}^{-1}\mathbf{f}$ because $\mathbf{L}$ is singular, we adopt the iterative solver in Poisson learning~\cite{calder2020poisson}. Specifically, we initialize $\mathbf{u}_0$ as an all-zero matrix in $\mathbb{R}^{n\times c}$ and writes the iteration step as
\begin{equation}
    \mathbf{u}_{t+1} \leftarrow \mathbf{u}_{t} + \mathbf{D}^{-1} (\mathbf{f}-\mathbf{L}\mathbf{u}_{t})
    \label{eq:iteration}
\end{equation}

The stopping criterion is determined by another loop $\mathbf{p}_{t+1}=\mathbf{W}\mathbf{D}^{-1}\mathbf{p}_t$, where $\mathbf{p}_0\in \mathbb{R}^{n}$ is initialized as a vector with ones at the positions of all labeled vertices and zeros elsewhere. Once $\|\mathbf{p}_t-\mathbf{p}_\infty\|_\infty\leq\frac1n$, where $\mathbf{p}_{\infty}=\mathbf{W}\mathbf{1}/(\mathbf{1}^\top \mathbf{W}\mathbf{1})$ represents the invariant distribution, the iteration is stopped. The number of iterations is denoted as $T$, which is often around 200-300.

We can unroll the iteration Eq.~\eqref{eq:iteration} and write $\mathbf{u} = \mathbf{u}_T$ in terms of $\mathbf{f}$ directly.
\[\mathbf{u}= \sum_{i=0}^{T-1} (\mathbf{D}^{-1}\mathbf{W})^i \mathbf{D}^{-1} \mathbf{f}:=\mathbf{A}\mathbf{f} \]

Then the objective function in optimization problem~\eqref{eq:optimization} can be written as
\[\argmin_{\mathbf{f}} ~ \frac{1}{m} \sum_{i=1}^{m} \|(\mathbf{A}\mathbf{f})_i-y_i\|_2^2 +\lambda \sum_{i=1}^{n}\|f_i\|_2^2\]

Since $f_i=0$ for $i \notin \mathcal{I}$, we extract the interface positions of $\mathbf{f}$ and denote them as $\mathbf{f}_\mathcal{I} \in \mathbb{R}^{|\mathcal{I}|\times c}$. $\mathbf{f}_\mathcal{I}$ is the interface term to be learned. Correspondingly, we extract the columns of $\mathbf{A}$ and denote them as $\mathbf{A}_\mathcal{I} \in \mathbb{R}^{n \times |\mathcal{I}|}$. Since the MSE loss only considers the $m$ labeled points, we can further extract the rows of $\mathbf{A}_\mathcal{I}$ that correspond to the $m$ training indices, denoted as $\tilde{\mathbf{A}}_\mathcal{I} \in \mathbb{R}^{m \times |\mathcal{I}|}$. This approach significantly reduces the space complexity because $m \ll n$. Finally, the optimization problem~\eqref{eq:optimization} becomes
\begin{equation}
    \argmin_{\mathbf{f}_\mathcal{I}} ~ \frac{1}{m} \left\|\tilde{\mathbf{A}}_\mathcal{I} \mathbf{f}_\mathcal{I} - \mathbf{y}\right\|_F^2 +\lambda \left\|\mathbf{f}_\mathcal{I}\right\|_F^2
    \label{eq:optimization_final}
\end{equation}

It is in fact the well-known $\ell^2$ regularized LS problem (a.k.a ridge regression) with $m$ observations and $|\mathcal{I}|$ variables. It has an explicit solution
\begin{equation}
    \mathbf{f}_{\mathcal{I}}^*= (\tilde{\mathbf{A}}_\mathcal{I}^\top \tilde{\mathbf{A}}_\mathcal{I}+m\lambda \mathbf{I})^{-1} \tilde{\mathbf{A}}_\mathcal{I}^\top \mathbf{y}
    \label{eq:explicit_solution}
\end{equation}

After we learn the interface term $\mathbf{f}_\mathcal{I}^*$, we can obtain the complete $\mathbf{f}^*$ by filling in the values of $\mathbf{f}_\mathcal{I}^*$ at the indices $i \in \mathcal{I}$, and setting the remaining elements to zero.

Notably, we employ the following methods to address complexity issues and solution non-uniqueness.
\begin{itemize}[leftmargin=*]
    \item In Eq.~\eqref{eq:explicit_solution}, we need to invert an $|\mathcal{I}| \times |\mathcal{I}|$ matrix, which is slow when $|\mathcal{I}|$ is large. A straightforward improvement comes from famous Sherman-Morrison-Woodbury formula, which performs inversion on $m\times m$ matrix instead.
    \begin{equation}
        \mathbf{f}_{\mathcal{I}}^*= \tilde{\mathbf{A}}_\mathcal{I}^\top ( \tilde{\mathbf{A}}_\mathcal{I}\tilde{\mathbf{A}}_\mathcal{I}^\top+m\lambda \mathbf{I})^{-1} \mathbf{y}
        \label{eq:explicit_solution_sherman}
    \end{equation}
    \item $\mathbf{L}\mathbf{u}=\mathbf{f}$ does not have a unique solution. Thus, we enforce a zero mean on each column of $\mathbf{u}$ by subtracting $\bar{\mathbf{u}}=\frac{1}{n}\sum_{i=1}^n u(x_i)$ along with each iteration step Eq.~\eqref{eq:iteration}. Consequently, the matrix $\mathbf{A}$ should be slightly modified as $\mathbf{A}= \sum_{i=0}^{T-1} \mathbf{J}(\mathbf{D}^{-1}\mathbf{W}\mathbf{J})^i \mathbf{D}^{-1}$, where $\mathbf{J}=\mathbf{I}-\frac{1}{n}\mathbf{1}\mathbf{1}^T$ is a projection matrix that removes the mean from each column.
\end{itemize}

\begin{algorithm}[t]
	\begin{algorithmic}[1]
	\INPUT $\mathbf{W}$, $\mathcal{L}_{\text{MSE}}^*$, $\mathbf{y}$, k-hop, iteration step $T$
    \PREPROCESS
	\State $\mathcal{I}$ = get\_interface\_idx (train\_idx, all\_idx, $\mathbf{W}$, k)
    \State Calculate $\mathbf{A}= \sum_{i=0}^{T-1} \mathbf{J}(\mathbf{D}^{-1}\mathbf{W}\mathbf{J})^i \mathbf{D}^{-1}$
    \State $\tilde{\mathbf{A}}_\mathcal{I}= \mathbf{A}[\text{train\_idx}, \mathcal{I}]$
    \State Solve $\lambda$ s.t. $g(\lambda)=\frac{1}{m} \left\|( \mathbf{I}+\frac{1}{m\lambda} \tilde{\mathbf{A}}_\mathcal{I} \tilde{\mathbf{A}}_\mathcal{I}^\top )^{-1} \mathbf{y}\right\|_F^2=\mathcal{L}_{\text{MSE}}^*$
    \TRAINING
	\State $\mathbf{f}_{\mathcal{I}}^*= \tilde{\mathbf{A}}_\mathcal{I}^\top ( \tilde{\mathbf{A}}_\mathcal{I}\tilde{\mathbf{A}}_\mathcal{I}^\top+m\lambda \mathbf{I})^{-1} \mathbf{y}$
    \State $\mathbf{f}^*=\mathbf{f}_{\mathcal{I}}^*$ if $i \in \mathcal{I}$;\quad $\mathbf{f}^*=0,$ otherwise.
    \INFERENCE
    \State $\mathbf{u}_{0}=\mathbf{0}$
    \For{$t=1,2,\cdots,T-1$}
	\State $\mathbf{u}_{t+1} \leftarrow \mathbf{u}_{t} + \mathbf{D}^{-1} (\mathbf{f}^*-\mathbf{L}\mathbf{u}_{t})$
    \State $\mathbf{u}_{t+1}=\mathbf{u}_{t+1}-\overline{\mathbf{u}_{t+1}}$
	\EndFor
    \State $\mathbf{u}=\mathbf{u}_{T}$
    \OUTPUT
    \State $\ell(x_i)=\underset{j\in\{1,\ldots,c\}}{\operatorname*{\operatorname*{\arg\max}}}\{u_j(x_i)\}.$
	\end{algorithmic}
	\caption{Interface Laplace Learning}
    \label{alg}
\end{algorithm}

Additionally, we provide a method for selecting the ridge parameter $\lambda$. By substituting $\mathbf{f}_\mathcal{I}$ in the optimization problem Eq.~\eqref{eq:optimization_final} with the closed-form solution $\mathbf{f}_{\mathcal{I}}^*$ given in Eq.~\eqref{eq:explicit_solution_sherman}, the first term, which corresponds to the MSE loss on labeled points, becomes
\[\frac{1}{m} \left\|\tilde{\mathbf{A}}_\mathcal{I} \mathbf{f}_\mathcal{I} - \mathbf{y}\right\|_F^2 =  \frac{1}{m} \left\|( \mathbf{I}+\frac{1}{m\lambda} \tilde{\mathbf{A}}_\mathcal{I} \tilde{\mathbf{A}}_\mathcal{I}^\top )^{-1} \mathbf{y}\right\|_F^2 := g(\lambda)\]
The function $g(\lambda)$, defined on $\lambda \in (0,\infty)$, is monotonically increasing and takes values in $(0,1)$. We observe that the final classification accuracy on unlabeled points depends on the resulting value of $g(\lambda)$. A detailed ablation study on the relationship between classification accuracy and the MSE loss is provided in~\Cref{subsubsec:ablation_mse}.

\begin{table*}[hbtp]
    \centering
    \caption{Average accuracy scores over 100 trials with standard deviation on MNIST, FashionMNIST and CIFAR-10.}
    \label{tab:main_results}
	\begin{tabular}{cllllll}
	\toprule
	\multicolumn{2}{c}{\# Label Per Class}& 1 & 2 & 3 & 4 & 5\\
	\midrule
	\multirow{11}{*}{MNIST} & Laplace~\cite{zhu2003semi} & 16.73 $\pm$ 7.41 & 28.04 $\pm$ 10.04 & 42.98 $\pm$ 12.18 & 54.90 $\pm$ 12.79 & 66.94 $\pm$ 12.06 \\
    & Nearest Neighbor & 55.24 $\pm$ 4.13 & 62.88 $\pm$ 3.02 & 67.31 $\pm$ 2.56 & 69.81 $\pm$ 2.37 & 71.39 $\pm$ 2.29 \\
    & Random Walk~\cite{zhou2004learning} & 83.12 $\pm$ 4.57 & 88.61 $\pm$ 2.12 & 91.18 $\pm$ 1.26 & 92.33 $\pm$ 0.98 & 93.06 $\pm$ 0.86 \\
    & MBO~\cite{garcia2014multiclass} & 13.03 $\pm$ 8.32 & 16.34 $\pm$ 9.37 & 21.23 $\pm$ 10.77 & 27.47 $\pm$ 10.50 & 33.62 $\pm$ 10.81 \\
    & WNLL~\cite{shi2017weighted} & 55.32 $\pm$ 13.61 & 84.86 $\pm$ 5.89 & 91.34 $\pm$ 2.80 & 93.68 $\pm$ 1.57 & 94.60 $\pm$ 1.17 \\
    & Centered Kernel~\cite{mai2018random} & 20.43 $\pm$ 2.18 & 25.94 $\pm$ 3.05 & 30.73 $\pm$ 3.52 & 34.63 $\pm$ 4.13 & 37.84 $\pm$ 4.07 \\
    & Sparse LP~\cite{jung2016semi} & 10.14 $\pm$ 0.13 & 10.14 $\pm$ 0.21 & 10.14 $\pm$ 0.22 & 10.18 $\pm$ 0.20 & 10.19 $\pm$ 0.22 \\
    & p-Laplace~\cite{rios2019algorithms} & 65.93 $\pm$ 4.89 & 75.72 $\pm$ 2.83 & 80.54 $\pm$ 1.99 & 83.02 $\pm$ 1.71 & 84.54 $\pm$ 1.56 \\
    & Poisson~\cite{calder2020poisson} & 90.58 $\pm$ 4.07 & 93.35 $\pm$ 1.64 & 94.47 $\pm$ 0.99 & 94.99 $\pm$ 0.65 & 95.29 $\pm$ 0.58 \\
    & V-Poisson~\cite{zhou2024variance} & 90.68 $\pm$ 4.89 & 93.98 $\pm$ 1.80 & 94.88 $\pm$ 0.94 & 95.20 $\pm$ 0.66 & 95.38 $\pm$ 0.56 \\
	& Inter-Laplace & \textbf{93.14} $\pm$ 3.82 & \textbf{95.21} $\pm$ 1.02 & \textbf{95.71} $\pm$ 0.64 & \textbf{95.93} $\pm$ 0.47 & \textbf{96.07} $\pm$ 0.40\\
	\midrule
	\multirow{11}{*}{FashionMNIST} & Laplace~\cite{zhu2003semi} & 18.77 $\pm$ 6.54 & 32.34 $\pm$ 8.98 & 43.44 $\pm$ 9.59 & 51.66 $\pm$ 7.50 & 57.38 $\pm$ 7.17 \\
    & Nearest Neighbor & 43.98 $\pm$ 4.87 & 49.51 $\pm$ 3.19 & 53.00 $\pm$ 2.57 & 55.20 $\pm$ 2.37 & 56.95 $\pm$ 2.15\\
    & Random Walk~\cite{zhou2004learning} & 55.43 $\pm$ 4.97 & 62.01 $\pm$ 3.20 & 66.00 $\pm$ 2.61 & 67.93 $\pm$ 2.45 & 69.64 $\pm$ 2.07 \\
    & MBO~\cite{garcia2014multiclass} & 11.27 $\pm$ 5.46 & 13.35 $\pm$ 6.24 & 15.76 $\pm$ 6.90 & 19.16 $\pm$ 7.85 & 22.63 $\pm$ 8.50 \\
    & WNLL~\cite{shi2017weighted} & 45.31 $\pm$ 7.08 & 59.24 $\pm$ 4.27 & 65.61 $\pm$ 3.32 & 68.30 $\pm$ 2.75 & 70.35 $\pm$ 2.40 \\
    & Centered Kernel~\cite{mai2018random} & 12.03 $\pm$ 0.37 & 13.35 $\pm$ 0.52 & 14.58 $\pm$ 0.81 & 15.95 $\pm$ 1.14 & 16.88 $\pm$ 1.05 \\
    & Sparse LP~\cite{jung2016semi} & 10.11 $\pm$ 0.18 & 10.17 $\pm$ 0.23 & 10.27 $\pm$ 0.25 & 10.26 $\pm$ 0.15 & 10.25 $\pm$ 0.17 \\
    & p-Laplace~\cite{rios2019algorithms} & 49.86 $\pm$ 5.13 & 56.91 $\pm$ 3.18 & 61.12 $\pm$ 2.48 & 63.46 $\pm$ 2.36 & 65.34 $\pm$ 2.03 \\
    & Poisson~\cite{calder2020poisson} & 60.13 $\pm$ 4.85 & 66.57 $\pm$ 3.07 & 69.97 $\pm$ 2.50 & 71.37 $\pm$ 2.20 & 72.67 $\pm$ 1.96 \\
    & V-Poisson~\cite{zhou2024variance} & 60.30 $\pm$ 5.64 & 66.70 $\pm$ 3.88 & 70.17 $\pm$ 2.97 & 71.44 $\pm$ 2.54 & 72.52 $\pm$ 2.14 \\
	& Inter-Laplace & \textbf{61.55} $\pm$ 5.08 & \textbf{68.02} $\pm$ 3.45 & \textbf{71.39} $\pm$ 2.65 & \textbf{72.72} $\pm$ 2.34 & \textbf{74.07} $\pm$ 1.92 \\
	\midrule
	\multirow{11}{*}{CIFAR-10} & Laplace~\cite{zhu2003semi} & 10.50 $\pm$ 1.35 & 11.27 $\pm$ 2.43 & 11.55 $\pm$ 2.62 & 12.78 $\pm$ 3.81 & 13.88 $\pm$ 4.59 \\
    & Nearest Neighbor & 30.06 $\pm$ 3.96 & 33.36 $\pm$ 2.95 & 35.21 $\pm$ 2.63 & 36.51 $\pm$ 2.32 & 37.70 $\pm$ 2.17 \\
    & Random Walk~\cite{zhou2004learning} & 38.97 $\pm$ 4.94 & 45.55 $\pm$ 3.70 & 49.15 $\pm$ 3.47 & 51.75 $\pm$ 2.98 & 53.48 $\pm$ 2.29 \\
    & MBO~\cite{garcia2014multiclass} & 11.07 $\pm$ 6.11 & 12.77 $\pm$ 6.76 & 14.01 $\pm$ 7.10 & 15.91 $\pm$ 7.15 & 17.43 $\pm$ 7.33 \\
    & WNLL~\cite{shi2017weighted} & 17.67 $\pm$ 5.58 & 27.28 $\pm$ 7.06 & 34.98 $\pm$ 6.80 & 40.52 $\pm$ 5.93 & 44.78 $\pm$ 4.69 \\
    & Centered Kernel~\cite{mai2018random} & 15.86 $\pm$ 1.81 & 17.71 $\pm$ 1.84 & 19.67 $\pm$ 2.15 & 21.53 $\pm$ 2.18 & 22.91 $\pm$ 2.42 \\
    & Sparse LP~\cite{jung2016semi} & 10.08 $\pm$ 0.10 & 10.07 $\pm$ 0.12 & 10.11 $\pm$ 0.22 & 10.03 $\pm$ 0.13 & 10.09 $\pm$ 0.15 \\
    & p-Laplace~\cite{rios2019algorithms} & 34.33 $\pm$ 4.65 & 40.33 $\pm$ 3.56 & 43.58 $\pm$ 3.10 & 45.99 $\pm$ 2.61 & 47.80 $\pm$ 2.17\\
    & Poisson~\cite{calder2020poisson} & 40.43 $\pm$ 5.48 & 46.63 $\pm$ 3.80 & 49.96 $\pm$ 3.84 & 52.39 $\pm$ 2.99 & 54.03 $\pm$ 2.35 \\
    & V-Poisson~\cite{zhou2024variance} & 34.40 $\pm$ 4.85 & 36.73 $\pm$ 4.02 & 37.65 $\pm$ 3.60 & 38.41 $\pm$ 3.67 & 39.07 $\pm$ 3.65 \\
	& Inter-Laplace & \textbf{41.74} $\pm$ 6.11 & \textbf{49.30} $\pm$ 4.13 & \textbf{53.46} $\pm$ 3.75 & \textbf{56.26} $\pm$ 3.25 & \textbf{58.21} $\pm$ 2.48\\
	\bottomrule
	\end{tabular}
\end{table*}

As a result, we propose to select a near-optimal $\lambda$ by solving the equation $g(\lambda) = \mathcal{L}_{\text{MSE}}^*$, where $\mathcal{L}_{\text{MSE}}^*$ is a dataset-specific target MSE value. Since $g(\lambda)$ is monotonic, its root can be efficiently computed using, for example, the bisection method. 

Although our derivation first obtains the closed-form solution $\mathbf{f}_{\mathcal{I}}^*$ of ridge regression and then substitutes it into the objective to define $g(\lambda)$, in practice this procedure can be viewed as a preprocessing step. Indeed, $\mathbf{f}_{\mathcal{I}}^*$ does not explicitly appear in the expression for $g(\lambda)$. In other words, we first determine $\lambda$ by solving $g(\lambda) = \mathcal{L}_{\text{MSE}}^*$, and then use this value of $\lambda$ to perform ridge regression and compute the solution $\mathbf{f}_{\mathcal{I}}^*$.

The final algorithm is provided in \Cref{alg}. Notice that we use the iterative solver during inference because we need the predictions $u(x_i)$ for all samples, rather than only for training samples as in the training stage.

\section{Experiments}
\subsection{Classification at very low label rates}

In this subsection, we conduct experiments to validate the effectiveness of our method under extreme label rates, with {1, 2, 3, 4, 5} labeled points per class, on the following real-world datasets: MNIST~\cite{lecun1998gradient}, FashionMNIST~\cite{xiao2017fashion}, and CIFAR-10~\cite{krizhevsky2009learning}. The MNIST and FashionMNIST datasets each contain 70,000 images, while CIFAR-10 contains 60,000 images, all collected from 10 distinct classes. Rather than using raw images to build the similarity graph, we follow the approach of~\cite{calder2020poisson}, which trains an autoencoder~\cite{kingma2022auto} to extract important features from the raw images. This generates a graph with higher quality for our method. The network architecture, loss function, and training procedure used for the autoencoder can be found in~\cite{calder2020poisson}. For fair comparison, we directly use the pre-computed features provided by~\cite{calder2020poisson} in our experiments.

After the features are extracted, we build a graph in the corresponding latent space. We use the Gaussian kernel (as defined in Eq.~\eqref{eq:gaussian_kernel}) to compute the edge weights between nodes in the graph. The pre-processing procedures used to construct the graph are exactly the same as those described in~\cite{calder2020poisson}: we set $\mathbf{W} = (\mathbf{W} + \mathbf{W}^\top) / 2$ for symmetry, and we set the diagonal entries of $\mathbf{W}$ to zero.

We compare our method with Laplace learning~\cite{zhu2003semi}, Random Walk~\cite{zhou2004learning}, multiclass MBO~\cite{garcia2014multiclass}, Weighted Nonlocal Laplacian (WNLL)~\cite{shi2017weighted}, Centered Kernel method~\cite{mai2018random}, Sparse Label Propagation~\cite{jung2016semi}, $p$-Laplace learning~\cite{rios2019algorithms}, Poisson learning~\cite{calder2020poisson} and Variance-enlarged Poisson learning~(V-Poisson)~\cite{zhou2024variance} in~\Cref{tab:main_results}. A nearest neighbor classifier, which decided the label according to the closest labeled vertex with respect to the graph geodesic distance, is provided as a baseline. The results for all methods, excluding V-Poisson, are obtained using the GraphLearning Python package \cite{graphlearning}. However, our implementation of the V-Poisson algorithm produces results that differ from the reported performance in \cite{zhou2024variance}. We discuss this discrepancy further in \Cref{sec:appendix_vpoisson}. In all experiments, we report the average accuracy and standard deviation across 100 random trials, with different labeled points selected each time. We test all methods on the same random permutations. Our method significantly outperforms others on various datasets. The code is publicly available at \url{https://github.com/shwangtangjun/Inter-Laplace}.

\paragraph{Time Complexity} The main computation burden is the calculation of the iteration matrix $\mathbf{A}$, which involves $T$ matrix multiplications between a dense $m \times n$ matrix and a sparse $n\times n$ matrix. Matrix multiplication is a highly optimized operation on GPUs. It takes approximately 0.4 seconds to compute $\mathbf{A}$ on a single NVIDIA GeForce RTX 3090Ti GPU. As observed in \Cref{tab:time_complexity}, the time consumed by other parts of the algorithm is negligible in comparison.

\begin{table}[hbtp]
    \centering
    \caption{Wall time elapsed (seconds) for each stage, evaluated on MNIST with 1 label per class. A single NVIDIA 3090Ti GPU is used.}
    \label{tab:time_complexity}
    \begin{adjustbox}{width = \columnwidth, center}
    \begin{tabular}{cccccc}
    \toprule
    \multicolumn{4}{c}{\textbf{Preprocess}} & \multicolumn{1}{|c|}{\textbf{Training}} & \textbf{Inference} \\
    get $T$ & get\_interface\_idx() & get $\mathbf{A}$ & get $\lambda$ & \multicolumn{1}{|c|}{get $\mathbf{f}^*$} & get $\mathbf{u}$ \\
    \midrule
    0.05 & 0.0065 & 0.4 & 0.008 & 0.0004 & 0.06 \\
    \bottomrule
    \end{tabular}
    \end{adjustbox}
\end{table}

\paragraph{Space Complexity} The storage of the $m\times n$ dense matrix $\tilde{A}$ accounts for the majority of the memory requirements. On a single NVIDIA GeForce RTX 3090Ti GPU with 24GB memory, our method can handle number of labeled samples $m$ up to 10,000.

\subsection{Ablation study on algorithm design}

There are several components that contribute to the overall performance of our method, such as the MSE objective function, squared $\ell^2$ regularization, interface positions approximation, and zero mean on each column. It is worthwhile to explore whether there exist superior alternatives to these components.


\subsubsection{Interface positions}
\label{subsec:interface_index}
In our algorithm, we remove k-hop neighbors of training samples to approximate the interface positions. To distinguish this method from others, we denote the resulting interface positions as $\mathcal{I}_{\text{khop}}$. To validate this choice, we test several other approaches: \textbf{(1) Ground truth}. Similar to the method of getting interface positions in~\Cref{subsec:interface_discontinuity}, we leak the label information and identify the indices of nonzero ground truth Laplacian values. \textbf{(2) Training}. $\mathcal{I}_{\text{training}}=\{1,2,\cdots,m\}$. \textbf{(3) All}. $\mathcal{I}_{\text{all}}=\{1,2,\cdots,n\}$. \textbf{(4) Random}. For fair comparison, the size of random indices is chosen to be equal to $|\mathcal{I}_{\text{khop}}|$. \textbf{(5) Laplace-base.} Firstly, we adopt Laplace learning~\cite{zhu2003semi} to get the prediction score $u(x_i)$ of each sample. This serves as a base method, not for direct classification, but for deciding the possible interface positions. We calculate the variance of each $u(x_i)$ and pick $|\mathcal{I}_{\text{khop}}|$ indices with the smallest variances. The base method is not limited to Laplace learning, as all classification methods can be possible options. We also test \textbf{Poisson-base} by adopting Poisson learning~\cite{calder2020poisson} as the base method. \textbf{(6) Geodesic.} We define the geodesic distance between two vertices as the sum of edge weights along the shortest path connecting them. we calculate the geodesic distance of other nodes to the training nodes using Dijkstra's algorithm, and choose the farthest $|\mathcal{I}_{\text{khop}}|$ nodes.

\begin{table}[hbtp]
    \centering
    \caption{Performance of different methods to approximate the interface positions with 1 label per class.}
    \label{tab:interface_index}
    \begin{tabular}{lccc}
    \toprule
    & MNIST & FashionMNIST & CIFAR-10 \\
    \midrule
    Remove k-hop & 93.14 $\pm$ 3.82 & 61.55 $\pm$ 5.08 & 41.74 $\pm$ 6.11 \\
    Ground truth & 90.33 $\pm$ 4.60 & 57.17 $\pm$ 5.19 & 38.45 $\pm$ 5.00 \\
    Training & 90.53 $\pm$ 4.14 & 60.13 $\pm$ 4.87 & 40.55 $\pm$ 5.17\\
    All & 90.38 $\pm$ 4.34 & 59.96 $\pm$ 5.17 & 39.41 $\pm$ 5.34 \\
    Random & 90.53 $\pm$ 4.46 & 59.89 $\pm$ 5.35 & 39.32 $\pm$ 5.43 \\
    Laplace-base & 93.08 $\pm$ 3.54 & 60.99 $\pm$ 5.22 & \textbf{41.84} $\pm$ 5.84\\
    Poisson-base & \textbf{93.62} $\pm$ 3.47 & \textbf{61.96} $\pm$ 5.88 & 41.69 $\pm$ 6.12\\
    Geodesic & 92.85 $\pm$ 3.87 & 61.34 $\pm$ 5.17 & 41.21 $\pm$ 6.18 \\
    \bottomrule
    \end{tabular}
\end{table}

We test each method on MNIST, FashionMNIST and CIFAR-10 with 1 labeled sample per class. The results are reported in \Cref{tab:interface_index}. Let's discuss each option one by one: Surprisingly, the so-called "ground truth" option, which leaks label information, does not perform well in real-world classification tasks. This indicates that finding interface positions for high-dimensional multi-class classification is much more complex than for synthetic data. Using the training indices, which is similar to Poisson learning~\cite{calder2020poisson}, indeed performs similarly to Poisson learning results in~\Cref{tab:main_results}. The key difference is that Poisson learning uses a fixed source term, while our approach tries to learn the interface term. However, their performance is inferior to the k-hop removal method, suggesting that the given labeled points should be treated as the interior rather than the boundary. Treating all samples as the interface is even worse than learning on a random subset. This indicates that the solution $\mathbf{u}$, while exhibiting discontinuity on the interface, should still maintain a certain level of smoothness in the interior. The Laplace-base and Poisson-base methods are useful as they further improve the classification accuracy by about 0.4\% on MNIST and FashionMNIST. This aligns with our intuition because the samples with the smallest prediction variance can be viewed as the hardest samples, and these are often located near the interface. However, in our proposed algorithm, we choose the k-hop removal approach because it is more efficient, easier to understand, and does not rely on other methods. The geodesic index approach, which uses a slightly more advanced version of graph distance, did not show improvements in the results. In conclusion, taking into account the implementation difficulty, classification performance, and execution efficiency, we choose to remove k-hop neighbors to approximate the interface positions.

\subsubsection{MSE loss}

\begin{table}[hbtp]
    \centering
    \caption{Comparison of Mean Squared Error and Cross-Entropy Loss performance.}
    \label{tab:ce_and_mse}
    \begin{tabular}{cccc}
        \toprule
        & \# Label & MSE Loss & CE Loss \\
        \midrule
        \multirow{5}{*}{MNIST} & 1 & 93.14 $\pm$ 3.82 & 93.01 $\pm$ 3.77 \\
        & 2 & 95.21 $\pm$ 1.02 & 95.10 $\pm$ 1.13\\
        & 3 & 95.71 $\pm$ 0.64 & 95.67 $\pm$ 0.68\\
        & 4 & 95.93 $\pm$ 0.47 & 95.89 $\pm$ 0.50\\
        & 5 & 96.07 $\pm$ 0.40 & 96.00 $\pm$ 0.38\\
        \midrule
        \multirow{5}{*}{FashionMNIST} & 1 & 61.55 $\pm$ 5.08 & 61.12 $\pm$ 4.97 \\
        & 2 & 68.02 $\pm$ 3.45 & 67.53 $\pm$ 3.54\\
        & 3 & 71.39 $\pm$ 2.65 & 71.17 $\pm$ 2.81\\
        & 4 & 72.72 $\pm$ 2.34 & 72.32 $\pm$ 2.53\\
        & 5 & 74.07 $\pm$ 1.92 & 73.89 $\pm$ 1.97\\
        \midrule
        \multirow{5}{*}{CIFAR-10} & 1 & 41.74 $\pm$ 6.11 & 40.43 $\pm$ 6.04 \\
        & 2 & 49.30 $\pm$ 4.13 & 48.41 $\pm$ 4.31\\
        & 3 & 53.46 $\pm$ 3.75 & 52.20 $\pm$ 3.90\\
        & 4 & 56.26 $\pm$ 3.25 & 54.93 $\pm$ 3.23\\
        & 5 & 58.21 $\pm$ 2.48 & 57.20 $\pm$ 2.52\\
        \bottomrule
    \end{tabular}
\end{table}

In multi-class classification problems, cross-entropy loss~(CE loss) is among one of the most popular choices of loss function. Specifically, the CE loss between the prediction $u(x_i)\in \mathbb{R}^c$ and integer-valued class label $y_i \in \mathbb{R}$ is defined as
\[\mathcal{L}_{\text{CE}}=-\frac{1}{m}\sum_{i=1}^m\log \frac{\exp(u(x_i)_{y_i})}{\sum_{j=1}^c \exp(u(x_i)_j)} \]
in which $u(x_i)_j$ denotes the $j$-th component of vector $u(x_i) \in \mathbb{R}^c$. The reason that we finally choose the MSE loss in our algorithm is two-fold:
\begin{itemize}[leftmargin=*]
    \item If we choose the CE loss, the optimization problem can no longer be formulated as a ridge regression problem and thus has no explicit solution. In this case, we use a gradient-based optimizer to minimize the loss function. Specifically, we adopt the L-BFGS method~\cite{liu1989limited} with learning rate 0.1, initializing $\mathbf{f}_{\mathcal{I}}$ by sampling from $\mathcal{U}(-0.5, 0.5)$. We find that 5 iterations are sufficient for convergence. Nonetheless, this approach requires approximately 0.4 seconds for training, which is about 1000$\times$ slower than computing the explicit solution to the LS problem under the MSE loss.
    \item The results using the MSE loss are better than using the CE loss, as shown in~\Cref{tab:ce_and_mse}. On MNIST and FashionMNIST datasets, the increase is marginal, but on CIFAR-10 we can see an increase of more than 1\%.
\end{itemize}

\subsubsection{Regularization norm}

The squared $\ell^2$ regularizer $\lambda |f|_2^2$ may be replaced by other regularization terms. Here, $f$ denotes the vector obtained by flattening $\mathbf{f}$, for consistency of notation in this subsection. We test three alternatives:
(1) $\ell^1$ norm: $|f|_1 = \sum |f_i|$;
(2) unsquared $\ell^2$ norm: $|f|_2 = \left(\sum f_i^2\right)^{1/2}$;
(3) $\ell^\infty$ norm: $|f|_\infty = \max |f_i|$.

Similar to the squared $\ell^2$ regularizer used in our method, these norms can also constrain the magnitude of the learned $f$. However, unlike ridge regression, optimization problems involving these norms generally do not admit explicit solutions. Moreover, since $\ell^p$ norms are non-differentiable at zero for $p \leq 1$ or $p = \infty$, gradient-based methods are not directly applicable to finding the corresponding minimizers.

\begin{table}[hbtp]
    \centering
    \caption{Comparison of different regularization norm on MNIST, 1 label per class.}
    \label{tab:l0l1l2}
    \begin{tabular}{cccc}
        \toprule
        $\|f\|_2^2$ & $\|f\|_1$ & $\|f\|_2$ & $\|f\|_\infty$ \\
        \midrule
        93.14 $\pm$ 3.82 & 89.93 $\pm$ 5.14 & 93.11 $\pm$ 3.69 & 92.72 $\pm$ 3.85 \\
        \bottomrule
    \end{tabular}
\end{table}

To solve this issue, we use FISTA~\cite{beck2009fast} (Fast Iterative Shrinkage Thresholding Algorithm) to handle the optimization problem. Briefly, the Iterative Shrinkage Thresholding Algorithm (ISTA) performs a gradient descent step on the non-regularized objective, followed by a shrinkage or projection step depending on the regularizer. Specifically, the $\ell^1$ norm corresponds to soft-thresholding, while the unsquared $\ell^2$ and $\ell^\infty$ norms correspond to projections onto an $\ell^2$ ball and an $\ell^1$ ball, respectively. FISTA improves upon ISTA by introducing Nesterov acceleration to speed up convergence.

We report the performance of different regularization norms on the MNIST dataset with one label per class in~\Cref{tab:l0l1l2}. It can be observed that the $\ell^1$ norm performs significantly worse than the others, while the unsquared $\ell^2$ and $\ell^\infty$ norms yield comparable results to the squared $\ell^2$ norm. However, due to the computational cost of the FISTA algorithm, which requires several seconds to converge (compared to an average of 0.0004 seconds for computing the closed-form solution with the squared $\ell^2$ regularizer), we adopt the latter in our method. Nonetheless, exploring the effect of different regularization norms remains an interesting direction for future work.


\subsubsection{Zero mean}

\begin{table}[hbtp]
    \centering
    \caption{Comparison of with~(w/) and without~(w/o) zero mean technique.}
    \label{tab:subtract_mean}
    \begin{tabular}{cccc}
        \toprule
        & \# Label & w/ & w/o \\
        \midrule
        \multirow{5}{*}{MNIST} & 1 & 93.14 $\pm$ 3.82 & 92.91 $\pm$ 3.89 \\
        & 2 & 95.21 $\pm$ 1.02 & 95.11 $\pm$ 1.11\\
        & 3 & 95.71 $\pm$ 0.64 & 95.66 $\pm$ 0.70\\
        & 4 & 95.93 $\pm$ 0.47 & 95.90 $\pm$ 0.52\\
        & 5 & 96.07 $\pm$ 0.40 & 96.03 $\pm$ 0.45\\
        \midrule
        \multirow{5}{*}{FashionMNIST} & 1 & 61.55 $\pm$ 5.08 & 60.83 $\pm$ 4.87 \\
        & 2 & 68.02 $\pm$ 3.45 & 67.58 $\pm$ 3.47\\
        & 3 & 71.39 $\pm$ 2.65 & 71.20 $\pm$ 2.81\\
        & 4 & 72.72 $\pm$ 2.34 & 72.50 $\pm$ 2.45\\
        & 5 & 74.07 $\pm$ 1.92 & 73.92 $\pm$ 1.98\\
        \midrule
        \multirow{5}{*}{CIFAR-10} & 1 & 41.74 $\pm$ 6.11 & 41.22 $\pm$ 6.13 \\
        & 2 & 49.30 $\pm$ 4.13 & 49.09 $\pm$ 4.30\\
        & 3 & 53.46 $\pm$ 3.75 & 53.20 $\pm$ 3.89\\
        & 4 & 56.26 $\pm$ 3.25 & 56.00 $\pm$ 3.19\\
        & 5 & 58.21 $\pm$ 2.48 & 58.02 $\pm$ 2.41\\
        \bottomrule
    \end{tabular}
\end{table}

In our algorithm, we add a mean subtraction step to ensure the uniqueness of solution, since the graph Laplacian matrix $\mathbf{L}$ is singular and admits infinite solutions. We manually choose the solution with zero mean along each component. To show that the effectiveness of our method does not come solely from extracting the mean, we report the results of our algorithm with and without mean subtraction step in~\Cref{tab:subtract_mean}. It is observed that the mean subtraction step helps increase the performance by less than 0.5\% on average. Notably, when we apply the same mean subtraction step to Poisson learning~\cite{calder2020poisson}, the performance of Poisson learning with and without this step shows very little difference, with less than 0.05\% difference.

\subsection{Ablation study on parameters}

There are two parameters in our experiment: k in k-hop and the target MSE loss $\mathcal{L}_{\text{MSE}}^*$. In~\Cref{tab:main_results}, we report the best results obtained through a grid search over the model parameters. The optimal k is provided in \Cref{tab:optimal_k}. We choose $\mathcal{L}_{\text{MSE}}^*=0.20$ for MNIST dataset, and $\mathcal{L}_{\text{MSE}}^*=0.35$ for FashionMNIST and CIFAR-10.

\begin{table}[hbtp]
	\centering
    \caption{Parameters k used to reproduce the results in \Cref{tab:main_results}.}
    \label{tab:optimal_k}
	\begin{tabular}{c@{\hskip7pt}ccccc}
	\toprule
	\# Label Per Class & 1 & 2 & 3 & 4 & 5\\
	\midrule
	MNIST & 4& 3 & 3 & 3 & 2 \\
	FashionMNIST & 5 & 4 & 4 & 3 & 3 \\
	CIFAR-10 & 3 & 3 & 2 & 2 & 2 \\
	\bottomrule
	\end{tabular}
\end{table}

In this subsection, we will study the effect of these two parameters.

\subsubsection{k-hop}

\begin{figure}[hbtp]
    \centering
    \subfloat[MNIST]{\includegraphics[width=0.5\linewidth]{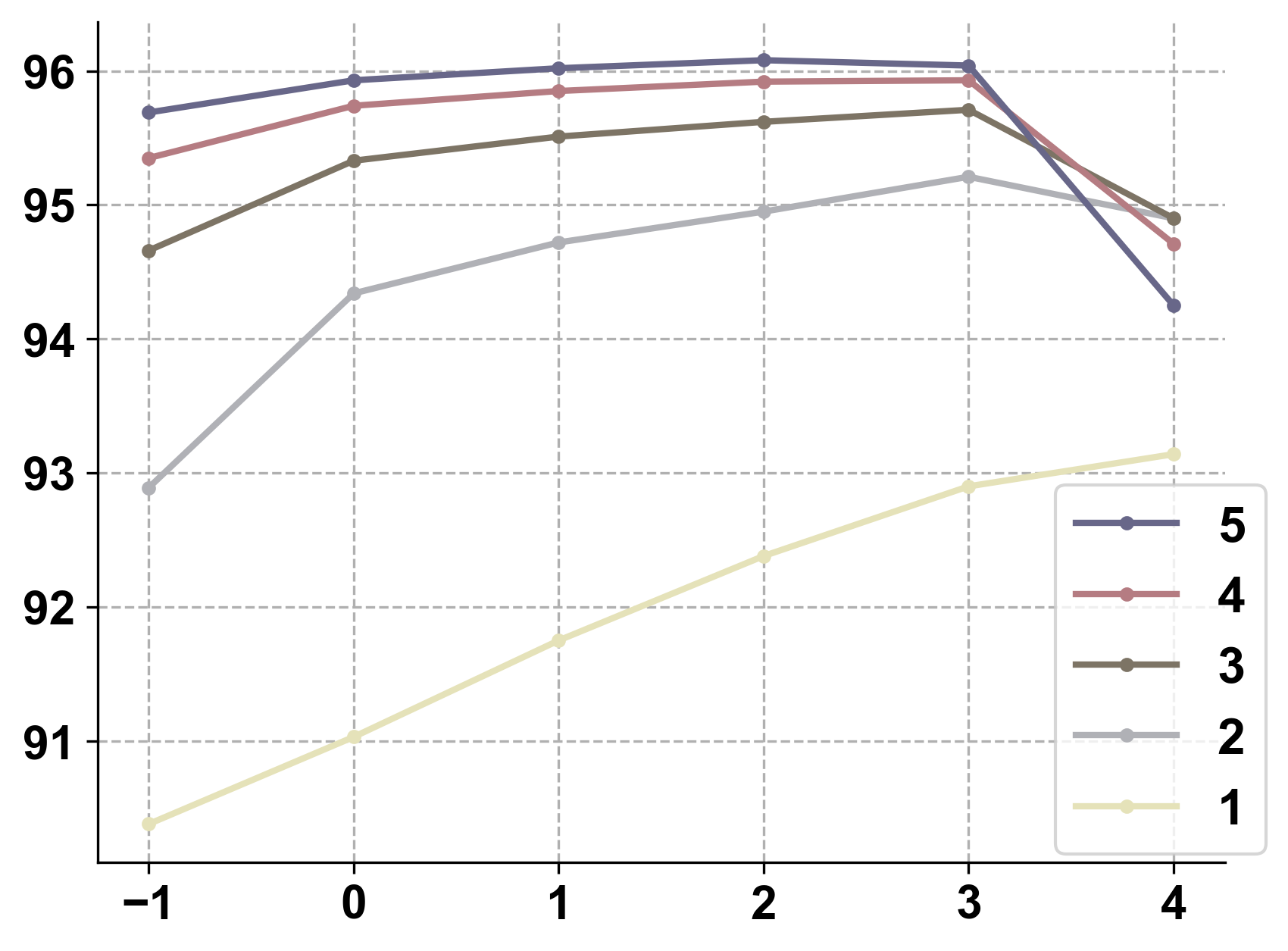}}
    \subfloat[CIFAR-10]{\includegraphics[width=0.5\linewidth]{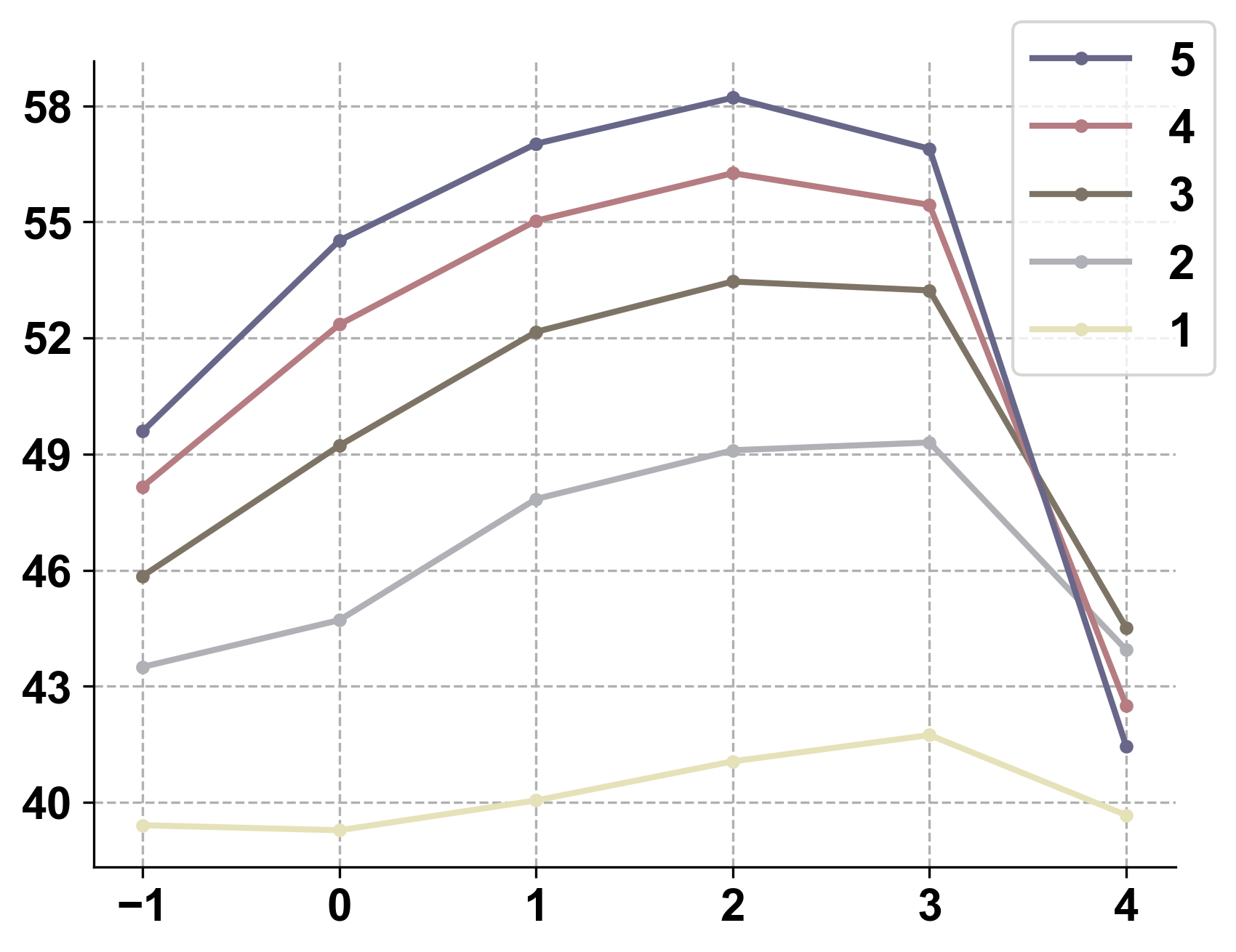}}
    \caption{Ablation study on k-hop parameter. x-axis: k-hop. y-axis: accuracy(\%). Each line corresponds to a different number of labeled samples per class.}
    \Description{Ablation study on k-hop parameter.}
    \label{fig:ablation_khop}
\end{figure}

We conduct experiments on MNIST and CIFAR-10, reporting the classification accuracy when k ranges from -1 to 4, and the number of labeled samples per class ranges from 1 to 5. Here, $\text{k}=-1$ means that all index are treated as the interface. For each value of k, we report the best performance with respect to the ridge regularization parameter $\lambda$. The results are presented in \Cref{fig:ablation_khop}. 

From the results, it is evident that removing k-hop indices is crucial for classification accuracy. The best accuracy at different label rates significantly outperforms the corresponding accuracy when $\text{k}=-1$. Moreover, we have more observations on the choice of k.
\begin{itemize}[leftmargin=*]
    \item When k is too large, the performance collapses. This is because the remaining positions for training the interface term are too small. For example, on the CIFAR-10 dataset, when the labeled number per class is 1 and $\text{k}=4$, there are only 2730/60000 points beyond the k-hop neighborhood. When the labeled number per class is 5 and $\text{k}=4$, there are only 44/60000 points beyond the k-hop neighborhood. With such a small number of trainable parameters, it may be too challenging for the algorithm to learn a good interface term that generalizes well.
    \item On the same dataset, the optimal k decreases as the number of labeled samples per class increases. On MNIST, when there is 1 label per class, the optimal $\text{k}=4$. When we increase the label number to 4, $\text{k}=3$ gives the best result. This phenomenon meets our expectation, as we want to keep the number of remaining indices moderate. The former setting gives 38018/70000 remaining indices, while the latter gives 42059/70000 remaining indices, which is comparable.
    \item For different datasets with the same number of labeled samples per class, the optimal k is different. This is because the connectivity of the datasets varies. Although the similarity graph W is constructed with the same number of nearest neighbors K=10, the underlying data distributions for different datasets are disparate. Hence, even with the same k, the remaining number of nonzero indices after removing the k-hop indices can be very different. For example, when the label number per class is 1 and $\text{k}=4$, there are 38018/70000 points left for MNIST, but only 2730/60000 points left for CIFAR-10. Such distinctions definitely lead to the different optimal k values for different datasets: the better connectivity of CIFAR-10 suggests that optimal k is smaller.
\end{itemize}

\subsubsection{$\mathcal{L}_{\text{MSE}}^*$}
\label{subsubsec:ablation_mse}


\begin{figure}[hbtp]
    \centering
    \subfloat[MNIST]{\includegraphics[width=0.5\linewidth]{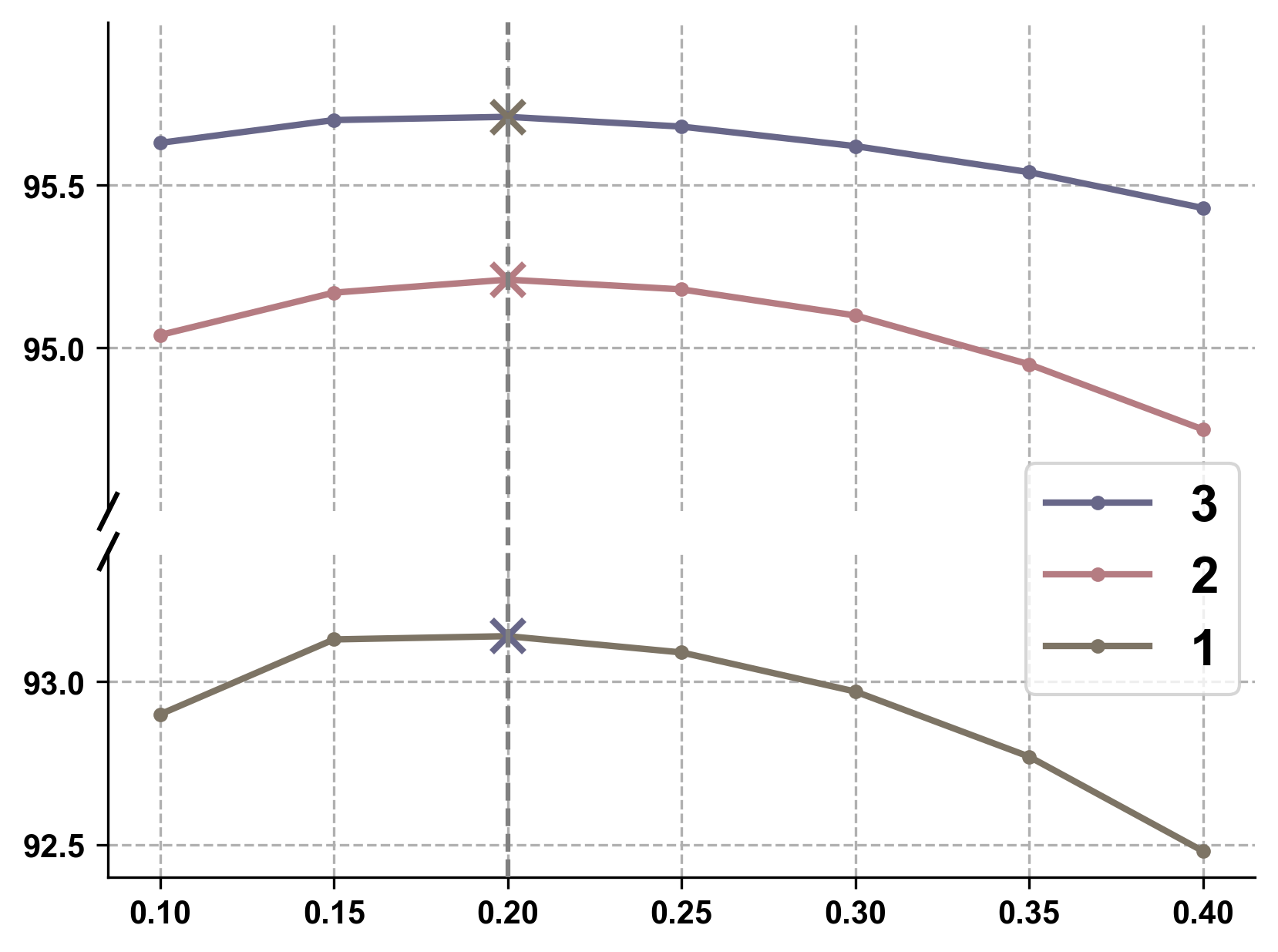}}
    \subfloat[CIFAR-10]{\includegraphics[width=0.5\linewidth]{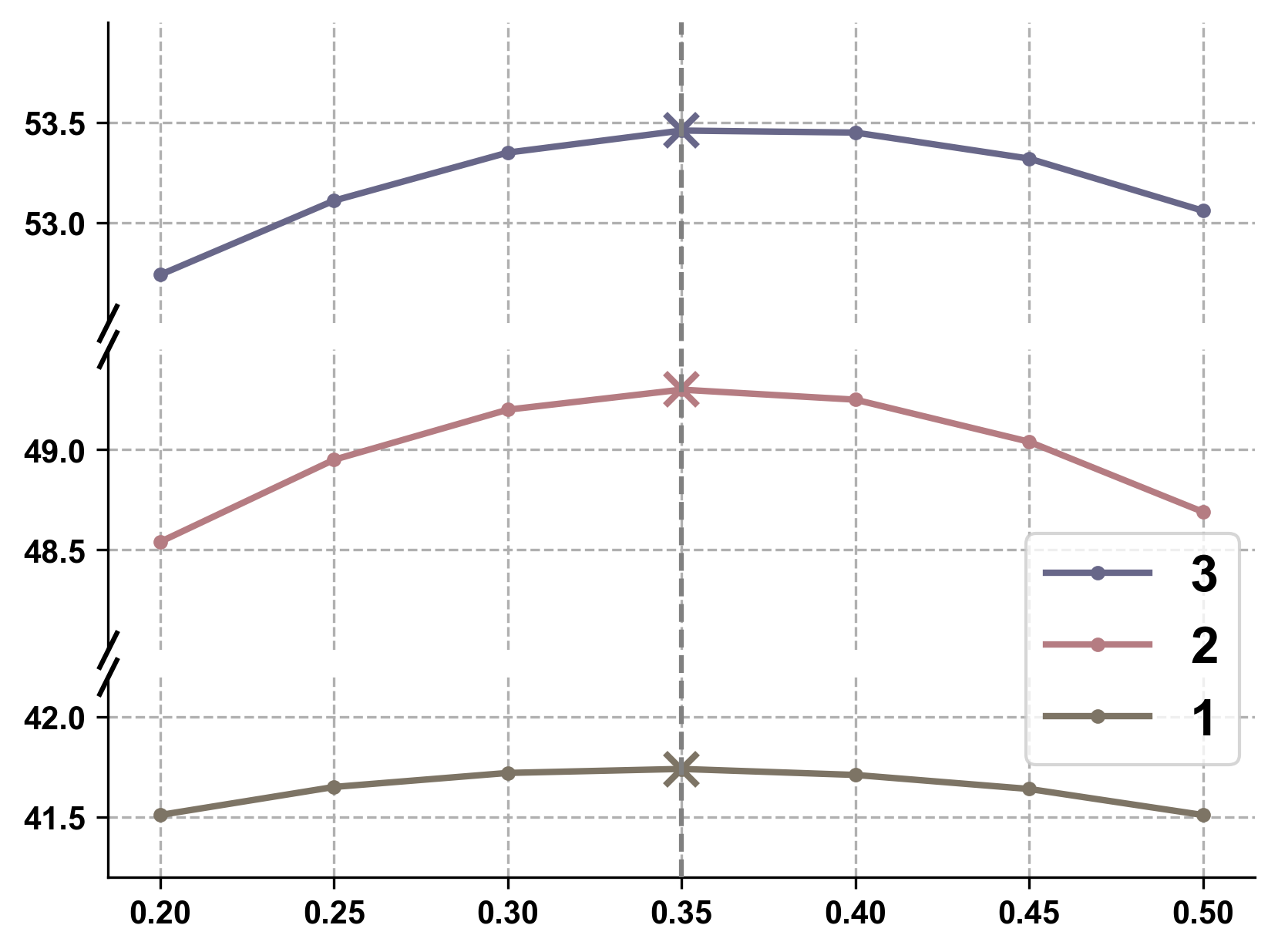}}
    \caption{Ablation study on $\mathcal{L}_{\text{MSE}}^*$. x-axis: $\mathcal{L}_{\text{MSE}}^*$. y-axis: accuracy(\%). Each line corresponds to a different number of labeled samples per class.}
    \label{fig:ablation_target_loss}
\end{figure}

Another important parameter is the target MSE loss $\mathcal{L}_{\text{MSE}}^*$, which guides the selection of the ridge parameter $\lambda$. In~\Cref{fig:ablation_target_loss}, we report the classification accuracy as a function of $\mathcal{L}_{\text{MSE}}^*$ under different numbers of labeled samples per class.

The results show that $\mathcal{L}_{\text{MSE}}^*$ is a dataset-specific parameter, in the sense that:
(1) for each dataset, the optimal $\mathcal{L}_{\text{MSE}}^*$ yielding the best accuracy remains roughly the same across different label rates;
(2) across different datasets, the optimal $\mathcal{L}_{\text{MSE}}^*$ varies—for example, approximately 0.20 for MNIST and 0.35 for CIFAR-10.

It is also noteworthy that the performance is fairly stable with respect to $\mathcal{L}_{\text{MSE}}^*$. For instance, on the CIFAR-10 dataset with one labeled sample per class, the accuracy remains above 41.5\% when $\mathcal{L}_{\text{MSE}}^*$ ranges from 0.20 to 0.50. Considering that the range of $\mathcal{L}_{\text{MSE}}^*$ is $(0,1)$, we conclude that the accuracy is robust to the choice of $\mathcal{L}_{\text{MSE}}^*$.

\subsection{Broader application scenarios}
In this subsection, we extend the application of our method to more scenarios, including cases with unbalanced label distribution and higher label rates per class.

\subsubsection{Unbalanced label distribution}

To account for the issue of unbalanced label distribution, we consider the following setup: for the even-numbered classes $\{0,2,4,6,8\}$, we use 1 labeled sample per class, while for the odd-numbered classes $\{1,3,5,7,9\}$, we use 5 labeled samples per class.

Other methods, such as Poisson Learning~\cite{calder2020poisson}, deal with unbalanced training data in a post-processing way by introducing a factor $s_{j}=b_{j}/n_{j}$ to the prediction $u(x_i)$:
$$\underset{j\in\{1,...,c\}}{\arg\max}\{s_ju_j(x_i)\}$$
where $b_j$ is the true proportion of class-$j$ samples in the dataset, and $n_j$ is the proportion of labeled points from class $j$ among all labeled samples. On the contrary, our method naturally learns a more suitable interface term from the unbalanced distribution during the training stage. We adopt the same weighting factor $s_j$, and apply it as the weight for the MSE loss in our objective function,
\[\frac{1}{m} \sum_{i=1}^{m} s_{y_i} \|(A\mathbf{f})_i-y_i\|_2^2 +\lambda \sum_{i=1}^{n}\|f_i\|_2^2\]
Here, $y_i$ in the expression $s_{y_i}$ refers to the integer-valued class label, not a one-hot encoded vector. We slightly abuse the notation for the sake of simplicity. This weighted loss approach is a widely applied technique when designing loss functions for handling unbalanced training data. Similar to \Cref{subsec:algorithm}, we can derive the explicit solution of the above objective function.
\[\mathbf{f}_{\mathcal{I}}^*= \tilde{\mathbf{A}}_\mathcal{I}^\top ( \tilde{\mathbf{A}}_\mathcal{I}\tilde{\mathbf{A}}_\mathcal{I}^\top +m \lambda \mathbf{S}^{-1})^{-1} \mathbf{y}\]
where $\mathbf{S}=\text{diag}(s_{y_i}) \in \mathbb{R}^{m\times m}$. The inference procedure remains unchanged from the previous algorithm, without the need for any additional post-processing steps.
\begin{table}[hbtp]
    \centering
    \caption{Performance of unbalanced label distribution: even classes 1 label per class, odd classes 5 labels per class.}
    \label{tab:unbalanced}
    \begin{tabular}{lccc}
    \toprule
    & MNIST & FashionMNIST & CIFAR-10 \\
    \midrule
     Poisson~\cite{calder2020poisson} & 93.88 $\pm$ 2.35 & 66.47 $\pm$ 3.80 & 46.87 $\pm$ 4.16 \\
     Inter-Laplace & 95.26 $\pm$ 2.07 & 68.16 $\pm$ 3.66 & 49.30 $\pm$ 4.72 \\
    \bottomrule
    \end{tabular}
\end{table}

The results are presented in \Cref{tab:unbalanced}. 
We compare our method to Poisson Learning, as it provides an approach to handle unbalanced training data. As a sanity check, both methods outperform their counterparts where only 1 label per class is provided, indicating that the additional labeled samples for the odd-numbered classes are indeed helpful. Importantly, our method outperforms Poisson Learning across all the datasets. This advantage stems from the fact that we incorporate the class imbalance directly into the objective function, allowing the interface term to learn from the unbalanced distribution, rather than relying on manual post-processing adjustments as in Poisson Learning.

\subsubsection{Higher label rate}

Although the motivation of interface La-place learning is to get accurate classification with low label rate, this approach remains effective in higher label rate scenarios. In experiments with 100 labeled samples per class, presented in~\Cref{tab:higher_label_rate}, our method outperforms other approaches across the datasets tested. While the gains are modest on simpler datasets like MNIST and FashionMNIST, the improvements become significant on more challenging dataset CIFAR-10. This result demonstrates the robustness of our approach in leveraging available labeled data effectively, regardless of the per-class label rate.

\begin{table}[hbtp]
    \centering
    \caption{Performance of higher label rate: 100 labels per class.}
    \label{tab:higher_label_rate}
    \begin{tabular}{l@{\hskip4pt}c@{\hskip7pt}c@{\hskip7pt}c}
    \toprule
    & MNIST & \makecell{Fashion-\\MNIST} & CIFAR-10 \\
    \midrule
    Laplace~\cite{zhu2003semi} & 96.83 $\pm$ 0.10 & 81.48 $\pm$ 0.39 & 63.88 $\pm$ 1.22 \\
    Nearest Neighbor & 85.58 $\pm$ 0.44 & 71.06 $\pm$ 0.55 & 49.32 $\pm$ 0.49 \\
    Random Walk~\cite{zhou2004learning} & 96.63 $\pm$ 0.11 & 81.37 $\pm$ 0.30 & 67.47 $\pm$ 0.51 \\
    MBO~\cite{garcia2014multiclass} & 96.88 $\pm$ 0.18 & 74.72 $\pm$ 0.91 & 42.40 $\pm$ 0.78 \\
    WNLL~\cite{shi2017weighted} & 96.25 $\pm$ 0.11 & 81.04 $\pm$ 0.30 & 67.04 $\pm$ 0.50\\
    Centered Kernel~\cite{mai2018random} & 88.66 $\pm$ 0.72 & 57.52 $\pm$ 2.24 & 57.30 $\pm$ 1.43 \\
    Sparse LP~\cite{jung2016semi} & 37.76 $\pm$ 1.12 & 28.12 $\pm$ 0.66 & 8.90 $\pm$ 0.22 \\
    p-Laplace~\cite{rios2019algorithms} & 93.54 $\pm$ 0.20 & 78.10 $\pm$ 0.39 & 63.46 $\pm$ 0.38 \\
    Poisson~\cite{calder2020poisson} & 96.77 $\pm$ 0.09 & 80.41 $\pm$ 0.62 & 66.09 $\pm$ 0.57 \\
    V-Poisson~\cite{zhou2024variance} & 95.89 $\pm$ 0.09 & 77.85 $\pm$ 0.52 & 41.67 $\pm$ 2.17\\
    Inter-Laplace & \textbf{97.25} $\pm$ 0.09 & \textbf{82.72} $\pm$ 0.28 & \textbf{73.28} $\pm$ 0.34 \\
    \bottomrule
    \end{tabular}
\end{table}

\section{Conclusion}
Inspired by the observation that interfaces exist between different classes, we propose a Laplace equation model with jump discontinuity and derive its nonlocal counterpart. Based on the nonlocal model, we introduce an interface term to enhance Laplace learning. We then design an effective algorithm to approximate the interface positions and learn the interface term. Experimental results verify that our method can accurately describe the data distribution and improve the performance of semi-supervised learning tasks. Our interface Laplace learning framework is general and may be extended to several important research areas such as node classification on graph-structured data and few-shot learning. Future work involves incorporating the interface concept into neural network architectures, such as graph neural networks, and exploring alternative approaches to approximate the interface position.


\bibliographystyle{ACM-Reference-Format}
\bibliography{reference}

\newpage
\appendix

\section{Other tries}
\subsection{Dirichlet-type interface term}
\begin{table}[hbtp]
    \centering
    \caption{Performance of Dirichlet-type interface term with 1 label per class}
    \label{tab:interface_laplace_dirichlet}
    \begin{tabular}{lccc}
    \toprule
    & MNIST & \makecell{Fashion-\\MNIST} & CIFAR-10 \\
    \midrule
    Laplace~\cite{zhu2003semi} & 16.73 $\pm$ 7.41 & 18.77 $\pm$ 6.54 & 10.50 $\pm$ 1.35 \\
    Poisson~\cite{calder2020poisson} & 90.58 $\pm$ 4.07 & 60.13 $\pm$ 4.85 & 40.43 $\pm$ 5.48 \\
    Inter-Laplace-D & 84.68 $\pm$ 4.06 & 59.18 $\pm$ 5.23 & 37.19 $\pm$ 4.44 \\
    Inter-Laplace & \textbf{93.13} $\pm$ 3.72 & \textbf{61.52} $\pm$ 5.04 & \textbf{41.71} $\pm$ 6.09 \\
    \bottomrule
    \end{tabular}
\end{table}
From the perspective of nonlocal models, we introduce the interface term $f_i$ through the graph Laplacian, i.e., $Lu(x_i)=f_i$. However, as Laplace learning~\cite{zhu2003semi} uses a Dirichlet-type boundary condition to incorporate the label information, it is natural to test whether our interface term can also be incorporated in a Dirichlet style. Specifically, we formulate the Dirichlet-type interface problem as the following:
\begin{alignat*}{2}
    Lu\left(x_i\right)&=0 && \quad i \notin \mathcal{I} \\
    u\left(x_i\right)&=f_i && \quad i \in \mathcal{I}
\end{alignat*}

Similar to~\Cref{subsec:algorithm}, we want to write $\mathbf{u}$ in terms of $\mathbf{f}$ in the form $\mathbf{u}=\mathbf{A}\mathbf{f}$. Denote $\mathcal{K}$ as the set of interior indices $ i \notin \mathcal{I}$. $\mathbf{L}_{\mathcal{K}} \in \mathbb{R}^{|\mathcal{K}|\times|\mathcal{K}|}$ extracts the corresponding rows and columns from the graph Laplacian matrix $\mathbf{L}$. $\mathbf{W}_{\mathcal{K},\mathcal{I}}\in \mathbb{R}^{|\mathcal{K}|\times |\mathcal{I}|}$ extracts $\mathcal{K}$ rows and $\mathcal{I}$ columns from the similarity matrix $\mathbf{W}$. Notice that unlike singular matrix $\mathbf{L}$, $\mathbf{L}_{\mathcal{K}}$ is invertible. Thus we can write the prediction $\mathbf{u}_{\mathcal{K}}$ on $i \in \mathcal{K}$ as:
\[\mathbf{u}_{\mathcal{K}}= \mathbf{L}_{\mathcal{K}}^{-1}\mathbf{W}_{\mathcal{K},\mathcal{I}} \mathbf{f}_{\mathcal{I}}\in \mathbb{R}^{|\mathcal{K}|\times c}  \]
The matrix inversion of $\mathbf{L}_{\mathcal{K}}$ is computationally expensive. However, as we only need the $m$ rows corresponding to the labeled indices in $\mathbf{u}_{\mathcal{K}}$ during training, we can instead solve the following equation
\[ \tilde{\mathbf{L}}_{\mathcal{K}}^{-1} \cdot \mathbf{L}_{\mathcal{K}} = \tilde{\mathbf{I}}\]
to calculate only the $m$ corresponding rows of $\mathbf{L}_{\mathcal{K}}^{-1}$. Here $\tilde{\mathbf{L}}_{\mathcal{K}}^{-1}$ and $\tilde{\mathbf{I}}$ indicates the rows that correspond to the $m$ training indices of $\mathbf{L}_{\mathcal{K}}^{-1}$ and identity matrix $\mathbf{I}$, respectively. Finally, we can write
\[\tilde{\mathbf{A}}_{\mathcal{I}}= \tilde{\mathbf{L}}_{\mathcal{K}}^{-1}\mathbf{W}_{\mathcal{K},\mathcal{I}}\]
We can then learn the interface term $\mathbf{f}^*$ in the same way as in \Cref{subsec:algorithm}. The algorithm with Dirichlet-type interface term is denoted as Inter-Laplace-D, and we compare its performance with Laplace learning, Poisson learning and our Inter-Laplace in \Cref{tab:interface_laplace_dirichlet}.

The proposed Inter-Laplace-D method can significantly improve performance compared to Laplace learning. This further substantiate our viewpoint that the function $u$ should exhibit discontinuities at category interfaces, while remaining smooth within their interiors.

However, the performance of Inter-Laplace-D does not quite match the results of Poisson learning and our Inter-Laplace method. The key difference is that Inter-Laplace imposes the interface term through $Lu(x_i)=f_i$, while Inter-Laplace-D imposes it through $u(x_i)=f_i$. We argue that the $Lu(x_i)=f_i$ constraint in Inter-Laplace is a more theoretically-principled approach, as it is derived from nonlocal models. Additionally, this formulation may be a smoother way to incorporate the required discontinuities. By restricting the second derivative $Lu(x_i)$, rather than just the function value $u(x_i)$, Inter-Laplace can more effectively capture the desired discontinuities at category boundaries. This distinction in interface term formulation may help explain why Poisson learning outperforms Laplace learning by a significant margin.

\subsection{Neural network parametrized interface term}
In our algorithm, we have directly treated the interface term $f_i$ as trainable parameters. However, an alternative approach could be to model $f_i$ as the output of a neural network. Specifically, we could construct a neural network with the extracted feature $x_i$ as input, and the network output $f_\theta(x_i)$ representing the interface term, where $f_\theta$ is the neural network with parameters $\theta$. This neural network-based formulation means that an explicit solution is no longer possible, as neural networks are inherently non-convex. Nonetheless, we can choose to optimize the same objective function as before:
\[ \argmin_\theta \frac{1}{m} \sum_{i=1}^{m} \|(A\mathbf{f}_\theta)_i-y_i\|_2^2 +\lambda \sum_{i=1}^{n}\|f_\theta(x_i)\|_2^2\]
where $\mathbf{f}_\theta=[f_\theta(x_1), \cdots, f_\theta(x_n)]^\top$. One may view the addition of neural networks as a way to incorporate feature information, since now the interface term $f_\theta(x_i)$ is dependent on the extracted feature $x_i$. This dependence on the input features can provide additional information to the learned interface term.

We conduct experiments on the MNIST dataset using 1 label per class. We construct a simple 2-layer MLP with a hidden dimension of 64 and ReLU activation function to parametrize $f_\theta$. The model is trained for 1000 epochs using the Adam optimizer with a learning rate of 0.01. Without incorporating feature information, our method achieves an average accuracy of 93.13 $\pm$ 3.72\%. However, when using a neural network approach, the performance drops to 92.19 $\pm$ 3.93\%. 

One possible explanation for this performance difference is that the feature information may already be sufficiently captured in the construction of the similarity matrix $W$. Incorporating the same information again through the neural network parametrization of $f_\theta$ could impose unnecessary restrictions on the learning process. Additionally, using a neural network introduces more hyper-parameters, such as the hidden dimension, choice of optimizer, learning rate, and network structure, which may need to be carefully tuned to achieve optimal performance.

Given the slightly worse performance observed when using a neural network to parametrize the interface term $f$, we opt not to use a neural network in our approach. However, we hypothesize that feature information might be helpful in graph node-classification tasks, such as the Cora~\cite{yang2016revisiting} dataset. In these tasks, the graph edge information is often considered orthogonal to the graph node information. Incorporating both the graph structure and the node features may lead to improved performance. We plan to explore this direction as part of our future work.

\begin{table*}[hbtp]
    \centering
    \caption{Average accuracy scores over 100 trials with standard deviation on \textit{new} MNIST. See \Cref{sec:appendix_vpoisson} for description for \textit{new} MNIST.}
    \label{tab:new_mnist}
    \begin{tabular}{llllll}
	\toprule
	\# Label Per Class & 1 & 2 & 3 & 4 & 5\\
    \midrule
    Laplace~\cite{zhu2003semi} & 17.74 $\pm$ 8.80 & 32.20 $\pm$ 12.23 & 49.47 $\pm$ 15.12 & 66.23 $\pm$ 12.80 & 76.45 $\pm$ 10.50 \\
    Nearest Neighbor & 57.50 $\pm$ 4.53 & 65.87 $\pm$ 3.22 & 70.49 $\pm$ 2.51 & 73.24 $\pm$ 2.43 & 74.98 $\pm$ 2.26\\
    Random Walk~\cite{zhou2004learning} & 85.00 $\pm$ 4.28 & 90.21 $\pm$ 2.13 & 92.53 $\pm$ 1.46 & 93.56 $\pm$ 1.23 & 94.24 $\pm$ 0.95 \\
    MBO~\cite{garcia2014multiclass} & 13.28 $\pm$ 8.66 & 17.10 $\pm$ 9.21 & 22.19 $\pm$ 10.77 & 28.01 $\pm$ 10.37 & 34.14 $\pm$ 11.67 \\
    WNLL~\cite{shi2017weighted} & 66.12 $\pm$ 14.03 & 90.51 $\pm$ 4.45 & 94.66 $\pm$ 1.54 & 95.77 $\pm$ 0.89 & 96.20 $\pm$ 0.50 \\
    Centered Kernel~\cite{mai2018random} & 19.52 $\pm$ 1.77 & 24.94 $\pm$ 2.70 & 29.86 $\pm$ 3.06 & 33.49 $\pm$ 3.22 & 36.72 $\pm$ 3.83\\
    Sparse LP~\cite{jung2016semi} & 10.02 $\pm$ 0.14 & 9.97 $\pm$ 0.22 & 10.00 $\pm$ 0.14 & 9.94 $\pm$ 0.14 & 9.79 $\pm$ 0.13 \\
    p-Laplace~\cite{rios2019algorithms} & 69.04 $\pm$ 4.79 & 78.56 $\pm$ 2.94 & 83.16 $\pm$ 2.25 & 85.58 $\pm$ 2.04 & 87.07 $\pm$ 1.80 \\
	Poisson~\cite{calder2020poisson} & 93.11 $\pm$ 3.87 & 95.20 $\pm$ 1.40 & 95.93 $\pm$ 0.71 & 96.22 $\pm$ 0.57 & 96.42 $\pm$ 0.35 \\
    V-Poisson~\cite{zhou2024variance} & 93.27 $\pm$ 4.35 & 95.49 $\pm$ 1.61 & 96.18 $\pm$ 0.48 & 96.31 $\pm$ 0.37 & 96.43 $\pm$ 0.38 \\
	Ours & \textbf{94.91} $\pm$ 3.83 & \textbf{96.34} $\pm$ 1.20 & \textbf{96.72} $\pm$ 0.46 & \textbf{96.80} $\pm$ 0.50 & \textbf{96.94} $\pm$ 0.36\\
    \bottomrule
    \end{tabular}
\end{table*}

\subsection{Synthetic regression}

\begin{figure}[hbtp]
    \centering
    \subfloat[Ground Truth]{\includegraphics[width=0.33\linewidth]{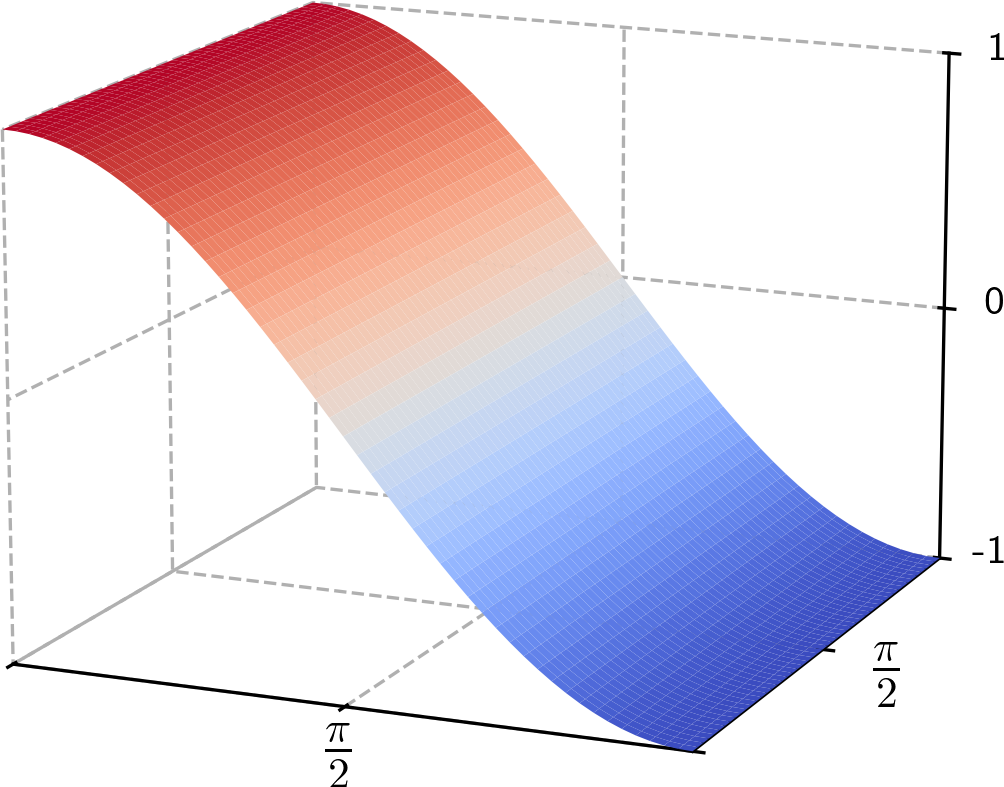}}
    
    \subfloat[Laplace. MSE=0.3118]{\includegraphics[width=0.33\linewidth]{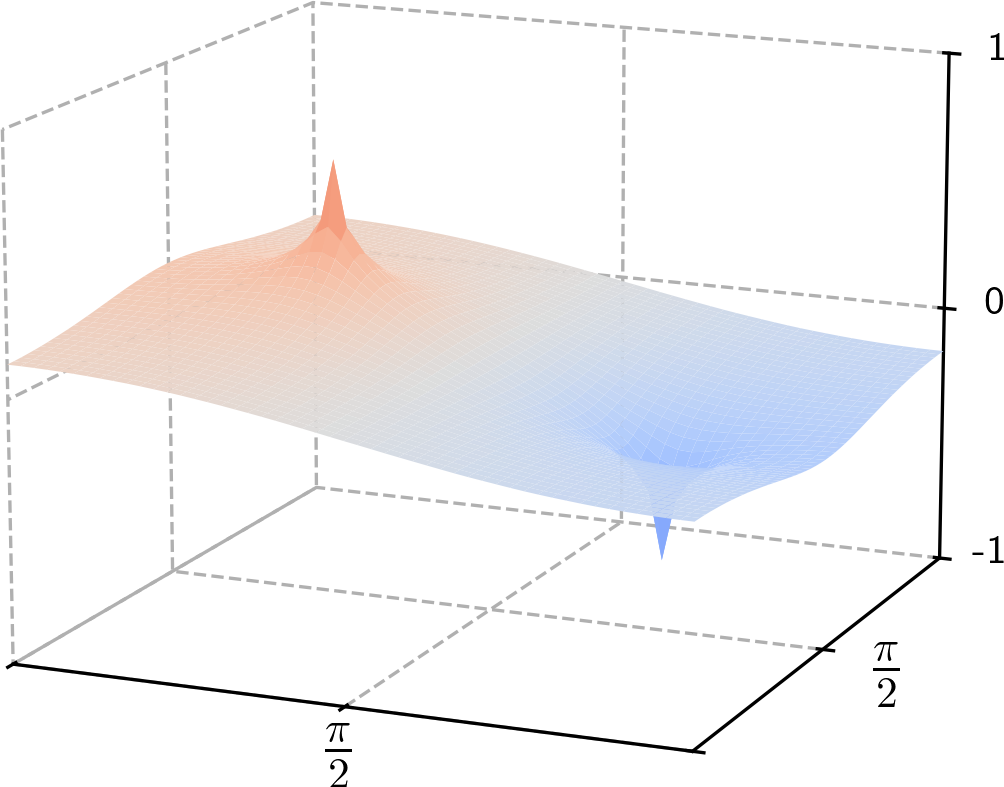}}
    \subfloat[Poisson. MSE=0.2598]{\includegraphics[width=0.33\linewidth]{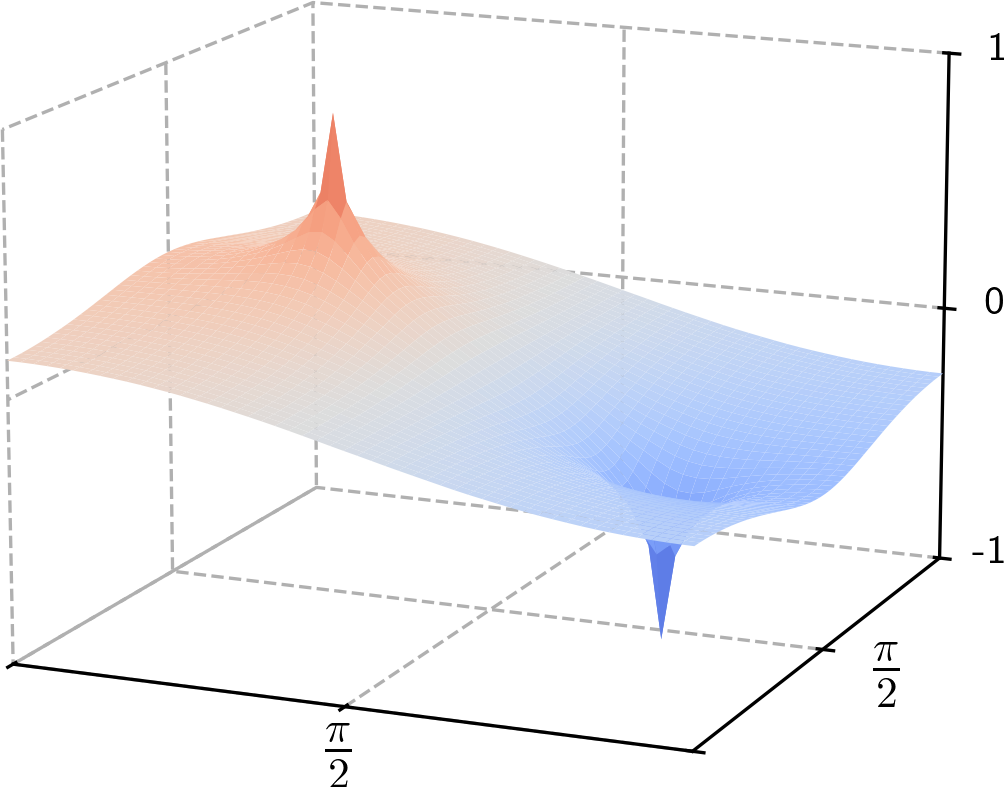}}
    \subfloat[Ours. MSE=0.0028]{\includegraphics[width=0.33\linewidth]{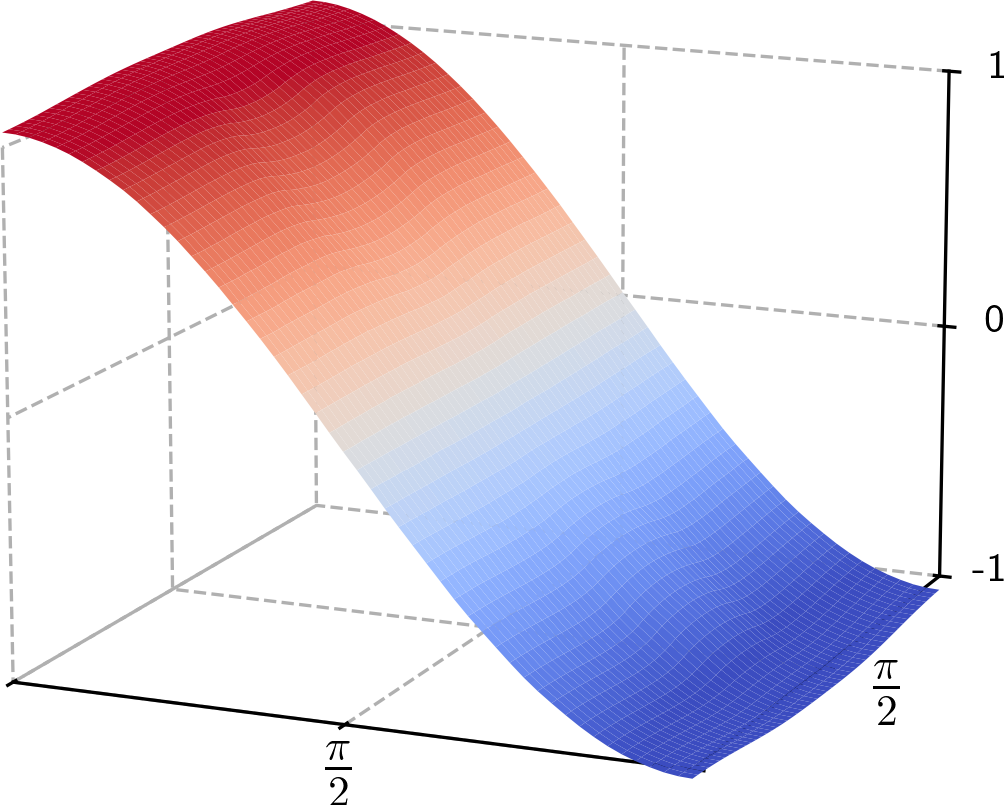}}
    \caption{A synthetic regression example.}
    \Description{A synthetic regression example.}
    \label{fig:regression}
\end{figure}

In~\Cref{subsec:interface_discontinuity}, we provide a toy classification example. It is noteworthy that our approach can also be applied to regression problems naturally by setting scalar values as the labels. In this subsection, we provide a comparison of regression results for Laplace learning, Poisson learning and our method on synthetic data.

We uniformly sample a regular $50 \times 50$ grid in $(x,y) \in [0,\pi]\times[0,\pi]$. The target is set as $\cos(x)$. We manually pick two labeled points, $(\frac{\pi}{4},\frac{\pi}{2})$ and $(\frac{3\pi}{4},\frac{\pi}{2})$. The similarity matrix is constructed as same as Eq.~\eqref{eq:gaussian_kernel}. For Poisson learning and our method, we use $T=10,000$ to ensure convergence. We pick k=5 and $\lambda=0.05$ in our method. The regression results and corresponding MSE are presented in~\Cref{fig:regression}. It is evident from the results that our method significantly outperforms the Laplace learning and Poisson learning approaches. While this particular regression problem does not exhibit a clear interface between distinct classes, it is still beneficial to assume that the underlying function is not harmonic almost everywhere.

\section{V-Poisson reproducibility issue}
\label{sec:appendix_vpoisson}

For the method Variance-enlarged Poisson learning (V-Poisson) proposed in~\cite{zhou2024variance}, we can only find a zip file on OpenReview, containing only partially reproducible code. The provided Github repository in the paper is empty. So we reproduce the results of V-Poisson by ourselves.  
\begin{itemize}[leftmargin=*]
    \item We use a different MNIST distance matrix from that used in V-Poisson. The authors of Poisson learning~\cite{calder2020poisson} provide a more fine-tuned distance matrix of MNIST in their GitHub repository two years after the original paper was published. However, since the exact training procedure for this updated distance matrix is not given, we decide to use the original distance matrix for experiments in the main paper. Nonetheless, we also provide the performance comparison on this updated "\textit{new} MNIST" distance matrix in \Cref{tab:new_mnist}. Our method still outperforms other approaches using this updated distance matrix. 
    \item The code for CIFAR-10 is missing in the OpenReview zip file mentioned earlier. The results reproduced by ourselves are not as good as those reported in ~\cite{zhou2024variance}. Missing of the original code makes it difficult to check the correctness of our reproduced results.
\end{itemize}
To clarify, we use the same off-the-shelf distance matrix for all methods in our experiments for a fair comparison . 

\section{Experiments on Self-Supervised Learned Representations on CIFAR-10.}

Recent self-supervised learning techniques based on contrastive learning have achieved impressive performance in learning representations, even in fully unlabeled settings. However, our focus lies in the extremely low-label regime, where fewer than 5 labeled samples per class are available. In such scenarios, deep self-supervised learning methods that rely on pretrained embeddings often struggle when directly applied to classification tasks. To illustrate this point, we conduct the following experiment using SimSiam~\cite{chen2021exploring}, a representative self-supervised model: (1) train a linear classifier directly on SimSiam features using 1 labeled sample per class (2) construct a similarity matrix $\mathbf{W}$ from SimSiam features and apply our Inter-Laplace algorithm.
\begin{table}[H]
	\centering
	\begin{tabular}{cc}
	\toprule
	Linear Classifier & Inter-Laplace \\
	\midrule
	37.73 $\pm$ 4.98 & 73.78 $\pm$ 7.31\\
	\bottomrule
	\end{tabular}
	\caption{Classification accuracy (\%) on CIFAR-10 with 1 labeled point per class, using SimSiam~\cite{chen2021exploring} feature. Mean accuracy and std among 100 random trails.}
	\label{tab:simsiam}
\end{table}
As shown in \Cref{tab:simsiam}, while SimSiam produces strong representations, the linear classifier performs poorly under scarce supervision. In contrast, our graph-based method significantly improves classification by effectively propagating the limited label information through the data graph. In one word, self-supervised models can provide good representations, but struggle to learn effective classifiers under extremly low label rate. Therefore, rather than positioning our method as a competitor to deep self-supervised learning techniques, we view it as complementary:
\begin{itemize}
	\item When strong representations are available (e.g., via contrastive learning), they can be used to construct better similarity graphs, thereby improving label propagation under our framework.
	\item Our method is orthogonal to how features are obtained — it can operate on raw distances, pretrained embeddings, or explicitly given graph structures.
\end{itemize}

\end{document}